\pdfoutput=1
\documentclass[11pt]{article}

\usepackage{ACL2023}

\usepackage{times}
\usepackage{latexsym}
\usepackage{xcolor}
\usepackage[T1]{fontenc}
\usepackage[utf8]{inputenc}

\usepackage{microtype}

\usepackage{inconsolata}

\usepackage{graphicx}
\usepackage{multirow}

\title{JMI at SemEval 2024 Task 3: Two-step approach for multimodal ECAC using in-context learning with GPT and instruction-tuned Llama models}

\author{Arefa$^{1, \dagger}$, Mohammed Abbas Ansari$^{1, \dagger}$, Chandni Saxena$^2$, Tanvir Ahmad$^1$\\
$^1$Jamia Millia Islamia University, New Delhi, India\\
$^2$The Chinese University of Hong Kong, Hong Kong SAR, China\\
{\tt \{arefa2001, mohd.abbas.ansari.2001\}@gmail.com}\\
{\tt csaxena@cse.cuhk.edu.hk, tahmad2@jmi.ac.in} \\
}
\begin{document}
\maketitle
\begingroup\def\thefootnote{$^{\dagger}$}\footnotetext{Equal contribution}\endgroup
\begin{abstract}This paper presents our system development for SemEval-2024 Task 3: "The Competition of Multimodal Emotion Cause Analysis in Conversations". Effectively capturing emotions in human conversations requires integrating multiple modalities such as text, audio, and video. However, the complexities of these diverse modalities pose challenges for developing an efficient multimodal emotion cause analysis (ECA) system. Our proposed approach addresses these challenges by a two-step framework. We adopt two different approaches in our implementation. In Approach~1, we employ instruction-tuning with two separate Llama 2 models for emotion and cause prediction. In Approach~2, we use GPT-4V for conversation-level video description and employ in-context learning with annotated conversation using GPT 3.5. Our system wins rank 4, and system ablation experiments demonstrate that our proposed solutions achieve significant performance gains. All the experimental codes are available on \href{https://github.com/CMOONCS/SemEval-2024_MultiModal_ECPE/tree/main}{Github}.
%Emotion Cause Analysis (ECA) is a prominent area within sentiment analysis that seeks to identify the causal factors behind emotions. It plays a crucial role in understanding the drivers of specific emotional responses. However, effectively capturing emotions requires leveraging various signals from human conversations, including text, audio, and video. Developing an efficient multimodal ECA system remains challenging due to the complexities among different modalities. 
\end{abstract}
\section{Introduction}
Emotion Cause Analysis (ECA) is centered around the extraction of potential cause clauses or pairs of emotion clauses and cause clauses from human communication, enabling a deeper understanding of communication dynamics. By incorporating multimodal cues like visual scenes, facial expressions, and vocal intonation, it facilitates a comprehensive and technically robust analysis of the factors that trigger diverse emotional reactions~\citep{mittal2021affect2mm,zhang2023cross,zheng2023facial}. Despite the considerable amount of research conducted using diverse audio, visual, and text modalities~\citep{gui2018event,xia2019emotion,fan2020transition,shoumy2020multimodal,abdullah2021multimodal}, there has been a noticeable gap in the exploration of multimodal ECA in natural settings (human conversations). In this context, \citet{DBLP:journals/taffco/WangDXLY23} introduce  Multimodal Emotion Cause Analysis in Conversations (ECAC) task and provide Emotion-Cause-in-Friends (ECF) dataset, which incorporates text, audio, and video modalities. This task consists of two sub-tasks: Textual Emotion-Cause Pair Extraction in Conversations (Subtask 1) and Multimodal Emotion Cause Analysis in Conversations (Subtask 2). A detailed description of these sub-tasks can be found in the task description paper~\citep{ECAC2024SemEval}. 

 %This paper presents our submission to Subtask~2 of multimodal ECAC, where we propose two distinct approaches to tackle the ECAC problem, both of which give competitive results. Inspired by the success of LLMs on various downstream tasks~\citep{wang2023augmenting,wang2024visionllm,Yang_2024_WACV}, especially emotion recognition, we devise two approaches using LLMs where we decompose the process of emotion-cause pair extraction into two steps. In the first step, we predict the emotions of the utterances in the conversation.
 In our submission to Subtask 2 of multimodal ECAC, this paper presents two distinct approaches to address the ECAC problem, giving competitive results. Drawing inspiration from the effectiveness of LLMs in diverse downstream tasks~\citep{wang2023augmenting,wang2024visionllm,Yang_2024_WACV}, including emotion recognition, we propose two LLM-based approaches that decompose the emotion-cause pair extraction process into two steps. The first step involves predicting the emotions of the utterances in the conversation. In the next step, we utilize these emotion labels to guide cause extraction. \textbf{Approach~1} involves instruction-tuning two separate Llama 2 models for emotion and cause prediction, while \textbf{Approach~2} leverages the in-context learning (ICL) capabilities~\citep{Dong2023ASF} of the GPT-3.5 model. Additionally, we introduce an efficient technique using the GPT-4V model to extract conversation-level descriptions from video modality.
 %In the next step, we use these emotion labels to guide the extraction of causes of emotional utterances. In our first approach, we instruction-tune two distinct Llama 2 models for emotion and cause prediction respectively. In the second approach, we use the in-context learning capabilities~\citep{Dong2023ASF} of the GPT-3.5 model for making predictions. We propose an efficient technique for extracting conversation-level descriptions from video modality using the GPT-4V model. 

 During the evaluation, our team ranked 4th on the leaderboard competing against more than 25 teams with a weighted-F1 score of 0.2816.
%The code is available on \href{https://github.com/CMOONCS/SemEval-2024_MultiModal_ECPE/tree/main}{Github}.

 %The extensive focus on transformer models and the rapid progress in LLMs has resulted in the development of models~\citep{imran2023uncovering, wang2023chatgpt} that demonstrate exceptional performance in general ECA tasks and datasets, including~\citep{lee2010emotion,poria2019emotion,poria2021recognizing}.

\section{Background}

\subsection{Task definition}
The input for the task, $D$, comprises \textit{N} conversations. As described by \citet{wang2021multimodal}, given a conversation $D_i=\{u_1, u_2, \dots, u_M\}$ consisting of $M$ utterances, where each utterance is represented by text, audio, and video, i.e. $u_j = [t_j, a_j, v_j]$, the goal of the task is to extract a set of emotion-cause pairs $P = \{\dots, (u_k^e, u_k^c), \dots\}$, where $u_k^e$ denotes an emotion utterance and $u_k^c$ corresponds to the cause utterance.

\subsection{Related Work}
The detailed Related Work section can be found in Appendix ~\ref{appendix:related_work}.

\subsection{Dataset}
\label{sec:appendixB}
We use the Emotion-Cause-in-Friends (ECF) dataset provided by \citet{DBLP:journals/taffco/WangDXLY23}, which is summarized in Table~\ref{tab:dataset-stats}. This dataset contains 13,509 multimodal utterances that occur in the American sitcom \textit{Friends} with 9272 emotion-cause pairs. Each utterance consists of the text, video, and audio.

\begin{figure}[htbp]
    \centering
    \includegraphics[width=\linewidth]{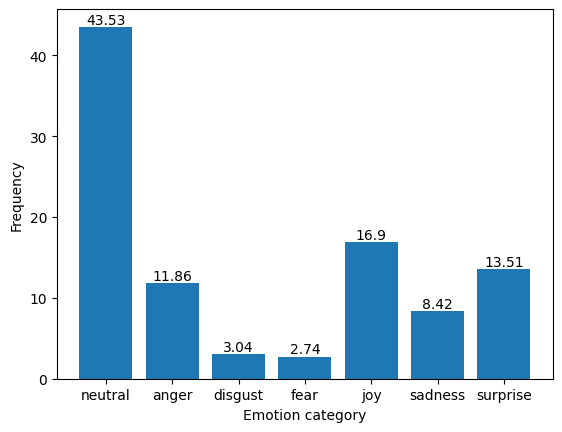}
    \caption{Percentage of each of the seven emotion categories}
    \label{fig:emotion-percentage}
\end{figure}

\paragraph{Class-distribution} {The dataset is imbalanced as shown in Fig.~\ref{fig:emotion-percentage} wherein around 44\% of the utterances have neutral emotion. Disgust and Fear constitute only 3\% and 2.7\% of the emotions.}

\begin{figure}[htbp]
    \centering
    \includegraphics[width=\linewidth]{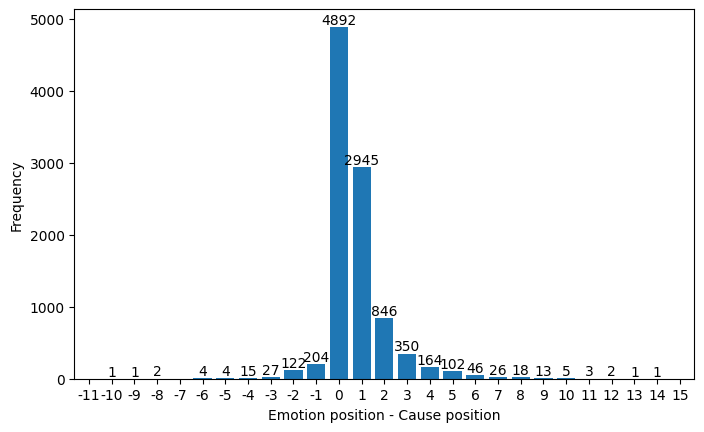}
    \caption{Relative position of emotion and causes}
    \label{fig:rel-pos}
\end{figure}

\paragraph{Relative positions of emotion and causes} {Interestingly, 49.95\% of the causes are self-causes meaning that the same utterance caused itself as shown in Fig.~\ref{fig:rel-pos}. This is also intuitive, as what one speaks or expresses often elicits the emotion of their utterance. Note that the dataset curators have also annotated utterances coming after the emotion utterance as its cause. These constitute only about 2.8\% of all causes and are one or two utterances away. 94.95\% of the causes are 0-5 utterances behind the emotion utterance. The fact that what you speak or other interlocutors in the conversation speak affects the emotion of subsequent utterances explains this phenomenon.}

\begin{table}
\centering
\small
{%
\begin{tabular}{lc}
\hline
\textbf{Items} & \textbf{Number} \\
\hline
Conversations & 1344    \\
Utterances    & 13,509      \\
Emotional Utterances & 7,690 \\
Self-Causal Utterances & 4,892 \\
Non-Self-Causal Utterances & 2,189 \\
No Cause Emotional Utterances & 609 \\
Later-Causal Utterances & 177 \\
\hline
\end{tabular}%
}
\caption{Statistics of causes for emotional utterances.}
\label{tab:dataset-stats}
\end{table}

%\paragraph{Related work} The detailed related work section can be found in the Appendix ~\ref{appendix:related_work}.

\section{Methodology}
\subsection{Overview}
{
We treat the task at hand as a two-step process. In the first step, we predict the emotion of each utterance in all \textit{N} conversations. Here, the context $C_j$ for utterance $u_j$ of conversation $D_i$ is the entire conversation itself. Given $E$ target emotion labels and $\hat{y_j^e}$ as the predicted emotion label, the problem can be formulated as (where $\theta$ denotes the parameters):
\begin{equation}
    \hat{y_j^e} = \arg \max_e \mathcal{P}(y^e | u_j, C_j, \theta)
\end{equation}

In the second step, given these emotion labels, we predict the causes of each utterance that has an emotion other than neutral. The causes will be a subset of all utterances in the conversation $D_i$. Let the learned function be $f: U \rightarrow 2^U$, where $U$ is the set of all utterances in the given conversation. It predicts the subset $\hat{y_j^c}$ of cause of emotion utterance $u_j$ where $\hat{y_j^e} \neq$ neutral as:
\begin{equation}
    \hat{y_j^c} = \arg \max_{y^c \in 2^U} \mathcal{P}(y^c | u_j, \hat{y_j^e}, C_j, \theta) 
\end{equation}
}

\subsection{Approach 1: Fine-tuned Llama-2}
{In our first approach, we perform instruction fine-tuning of the Llama 2 Large Language Model, an open-source model developed by GenAI, Meta \citep{touvron2023llama}. From the three variants with 7, 13, and 70 billion parameters, we use the 13 billion parameter model due to resource constraints, albeit the performance of this model achieves state-of-the-art results on various downstream NLP tasks compared to other models of similar sizes \citep{touvron2023llama}. In addition, we use the Llama 2-chat version of the model\footnote{https://huggingface.co/meta-llama/Llama-2-13b-chat-hf}, which is optimized for dialogue use cases as it aligns with our task.
%We started with prompt engineering where we used zero-shot prompting on the LLM using the API\footnote{https://www.llama2.ai/} to choose the prompts that gave the best results for emotion as well as cause prediction. 
In our approach, we use Llama2 API\footnote{https://www.llama2.ai/} for prompt engineering. Through zero-shot prompting, we select optimal prompts for emotion identification and cause prediction.
%We found that the output of the model was better when we treated the two tasks as separate, meaning we first find the emotions of all utterances in the conversation. Then, we add these labels to the conversation and prompt the model to predict the causes for each emotion utterance. Hence, we perform supervised fine-tuning of two separate Llama 2 models for these two tasks. While this increases the inference time, the performance gains far outweigh this latency introduced. We treat both these tasks as conditional generation where given the prompt, the model outputs the emotion label in the first case and the cause list in the second case. Detailed descriptions are given in the next two sections. Figure \ref{fig:approach1} shows the fine-tuning procedure.
We observed that treating these two tasks separately resulted in better model output. This approach involves first identifying the emotions of all utterances in the conversation. We then add these emotion labels to the conversation and prompt the model to predict the causes for each emotion utterance. Consequently, we perform supervised fine-tuning of two separate Llama 2 models for these tasks. Although this increases the inference time, the significant performance gains outweigh the introduced latency. We treat both tasks as conditional generation, where the model generated the emotion label in the first case and the cause list in the second case, given the prompt. Detailed explanations of these approaches are provided in the following sections. The fine-tuning procedure is shown in Fig.\ref{fig:approach1}. 
\begin{figure*}
    \centering
    \includegraphics[width=\textwidth]{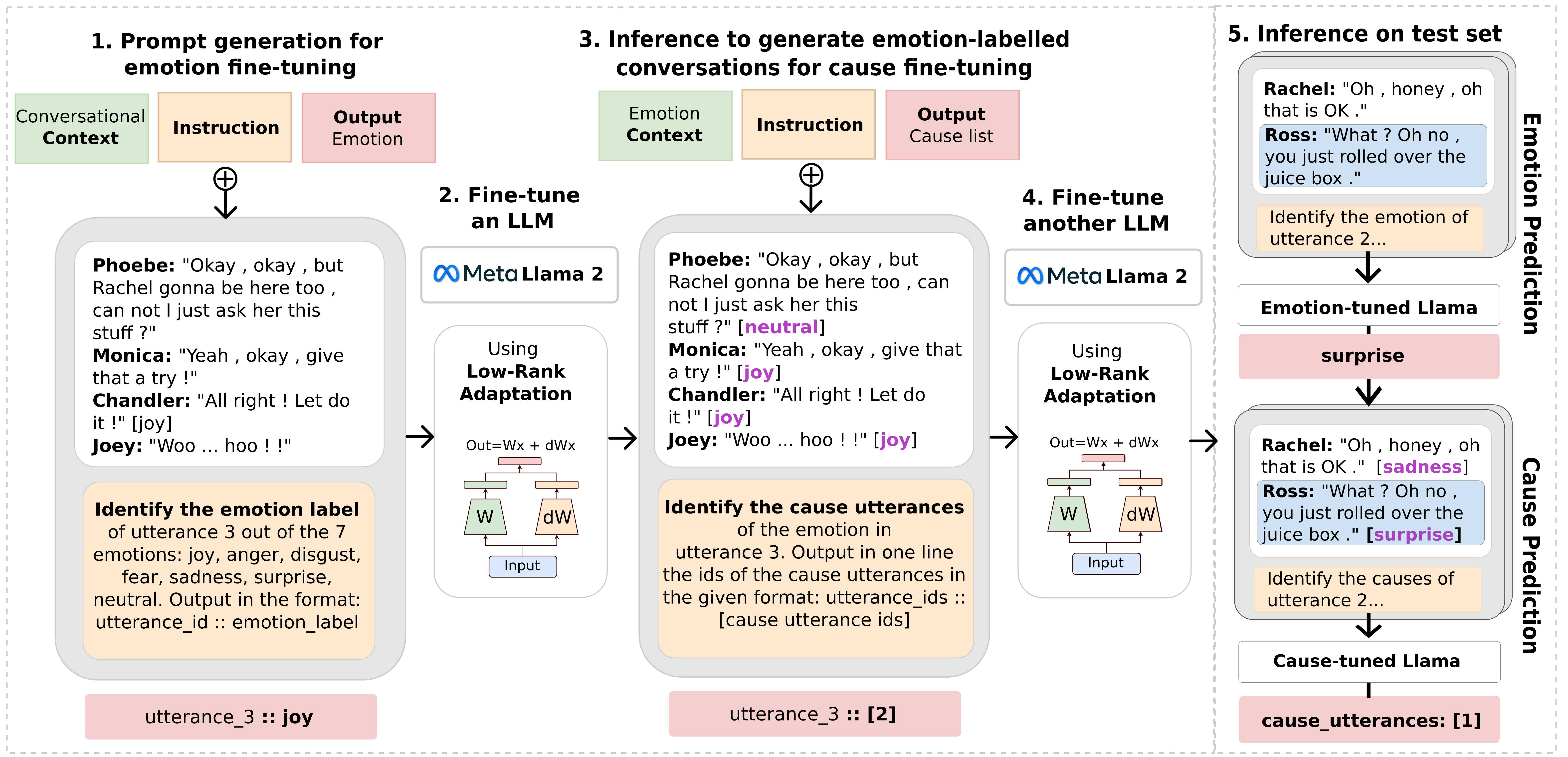}
    \caption{Pipeline for fine-tuning Llama (Approach 1)}
    \label{fig:approach1}
\end{figure*}

\subsubsection{Emotion recognition}
%For performing emotion recognition, we start by creating the dataset where each sample consists of an utterance $u_j$ from one of the $N$ conversations $D$ for which the LLM needs to output the emotion label.
To perform emotion recognition, we create a dataset where each sample includes an utterance $u_j$ from one of the $N$ conversations $D$ for which the LLM needs to output the emotion label.
%We then add the entire conversation $D_i$ along with speaker information as context to our prompt. 
We incorporate the entire conversation $D_i$ along with speaker information as context in our prompt. %We add this contextual information to boost the model's ability to understand the flow of emotions within the entire conversation for making its prediction as shown by our ablation studies in Section \ref{sec:ablation}.
This contextual information enhances the model's understanding of the flow of emotions within the conversation, as demonstrated by our ablation studies in Section~\ref{sec:ablation}. The instruction $I_j^e$, which gave the best results, is given in Appendix ~\ref{appendix:llama-prompts} along with detailed prompt examples. The prompt consists of the instruction $I_j^e$ and the context $C_j$ for utterance $u_j$:
\begin{equation}
    Prompt_j = (C_j, I_j^e)
\end{equation}
Using this prompt as the input and the corresponding true emotion label $y_j^e$, we perform supervised fine-tuning of a Llama 2-13b model. 
\begin{equation}
    \hat{y_j^e} = \textbf{Llama}_e(Prompt_j, \theta)
\end{equation}

We use a quantized version of the model due to memory limitations and perform Quantized Low-Rank Adaptation (QLoRA)~\citep{dettmers2024qlora} as a parameter-efficient fine-tuning technique. The training details are provided in the Section~\ref{section:training}. 

\subsubsection{Cause prediction}
%After obtaining the emotion labels for each utterance, we now create the dataset for cause prediction. 
To prepare the dataset for cause prediction, we incorporate the emotion labels obtained for each utterance. %Here, the context, i.e. the conversation, also consists of the emotion labels for each utterance $u_j$ where the predicted emotion label $\hat{y_j^e}$ is not \textit{neutral}. 
The conversation context now includes the emotion labels for each utterance $u_j$ excluding those with a predicted emotion label $\hat{y_j^e}$ of \textit{neutral}. This approach enhances the model's ability to analyze causal dependencies and identify which utterances may have contributed to a specific emotion. The output for cause prediction is a list of cause utterance IDs. The instruction is provided in Appendix ~\ref{appendix:llama-prompts}. The modified prompt for this step consists of this instruction $I_j^c$ along with the conversational context with emotion labels $C_{j}^e$:
\begin{equation}
    Prompt_j = (C_j^e, I_j^c)
\end{equation}
Next, we perform supervised fine-tuning of a new Llama 2-13b model using this prompt as the input and the corresponding true list of causes:
\begin{equation}
    \hat{y_j^c} = \textbf{Llama}_c(Prompt_j, \theta)
\end{equation}

\subsubsection{Adding video captions}
To integrate cues from the videos corresponding to each utterance, we experimented using video captions generated using GPT-4 Vision as additional context for the model. However, we observed a notable decrease in performance since descriptions for individual utterances were somewhat noisy and did not effectively guide the predictions. Moreover, the captions often contained multiple emotions causing confusion for the model. As a result, we do not utilize these during training.
%since the descriptions for individual utterances were a bit noisy and did not help guide the predictions. Often, they would contain multiple emotions, confusing the model. Hence, they are not used.
 
\subsection{Approach 2: In-Context-Learning GPT}
\begin{figure*}[t]
    \centering
    \includegraphics[width=\linewidth]{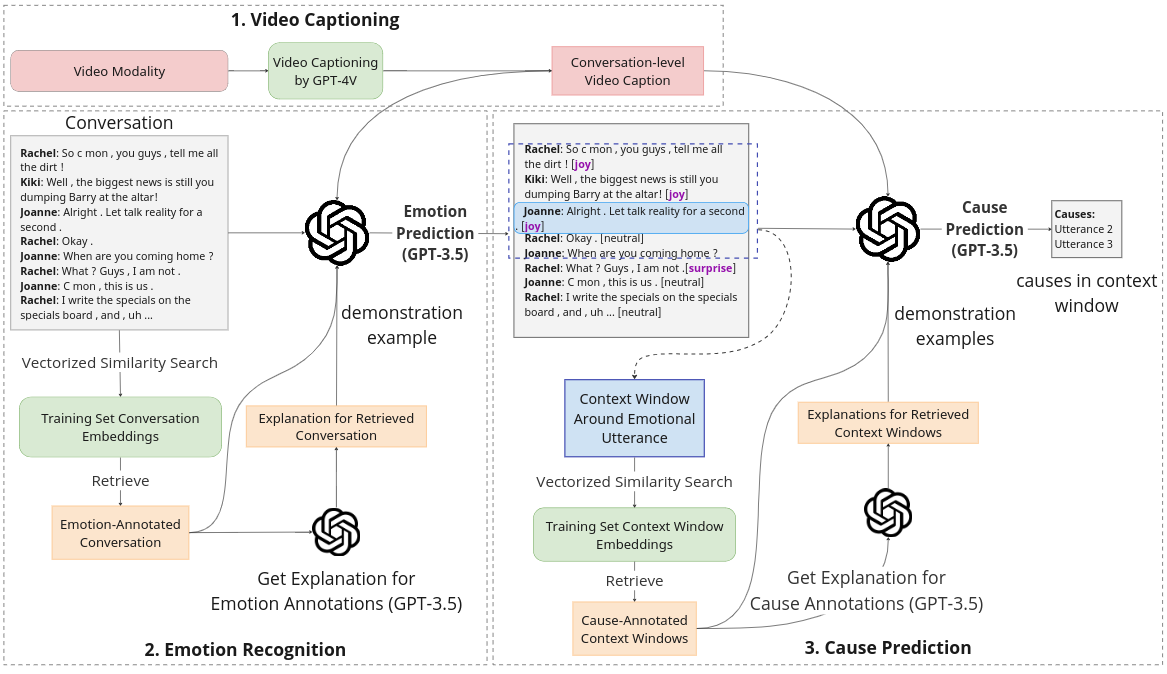}
    \caption{Pipeline of In-Context-Learning GPT Method (Approach 2)}
    \label{fig:gpt-pipeline}
\end{figure*}

Our second approach (Fig.~\ref{fig:gpt-pipeline}) tackles subtask 2 by obtaining conversation-level video captions using the GPT-4V(ision) model by OpenAI \cite{Yang2023lmms}. For emotion prediction, we retrieve a semantically similar conversation from the training set whose emotion annotations are explained as demonstration examples in the prompt for the GPT-3.5 model\footnote{https://platform.openai.com/docs/models/gpt-3-5-turbo}. For each predicted emotional utterance, we perform cause prediction within a context window around the emotional utterance. Due to the complex nature of the task, we leverage in-context-learning \cite{Dong2023ASF} by retrieving similar context windows from the training set whose cause annotations are explained as demonstration examples in the prompt for the GPT-3.5 model. We discuss each step in the subsequent sections.
\subsubsection{Video Captioning}

GPT-4V has the capability to process video sequences \citep{Yang2023lmms,Lin2023MMVIDAV}. In our approach, we extract conversation-level captions from the videos. However, due to rate limits and the costs considerations, we use a compact image representation for each video associated with the utterances of a conversation. %we tackle this task by constructing a compact image representation of each video associated with the utterances of a conversation. Thus, we give as input the sequence of images corresponding to utterances to the GPT-4V model to generate a description of the whole conversation. 
Therefore, these image sequences serve as input to the GPT-4V model, generating a description for the entire conversation. The prompt is shown in the Fig.~\ref{fig:video-caption}.

\begin{figure}[t]
    \centering
    \includegraphics[width=\linewidth]{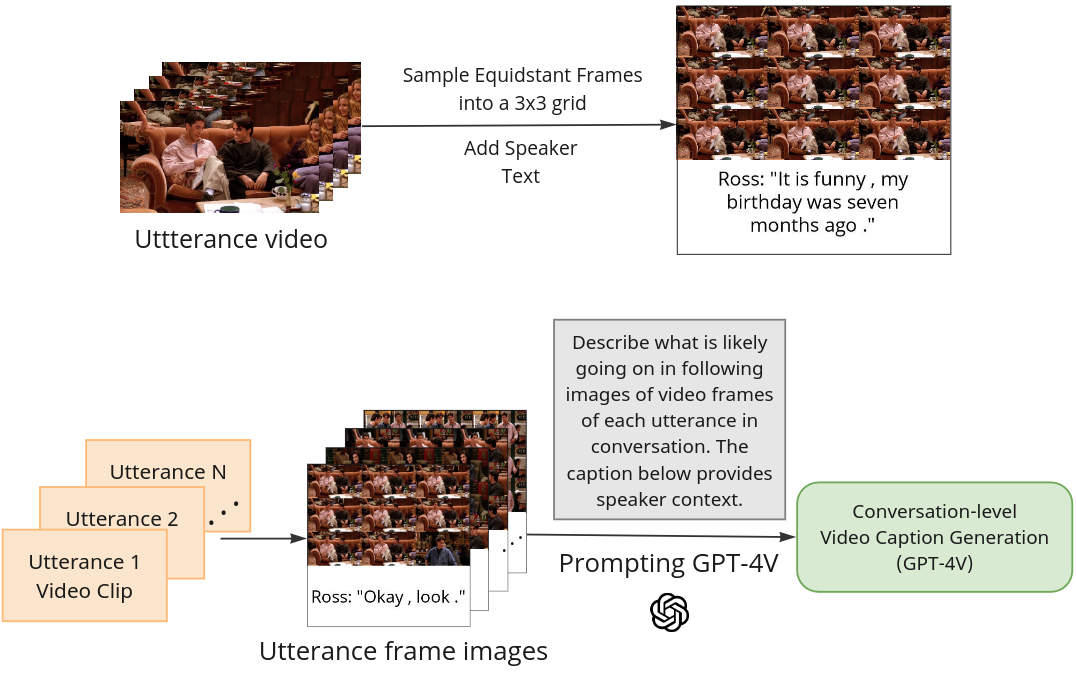}
    \caption{Video Captioning Pipeline}
    \label{fig:video-caption}
\end{figure}

For an utterance, we sample nine equidistant frames across its video length. These frames aim to capture the dynamics of the whole video. We arrange these frames in a $3\times3$ grid, following a row-major order. Additionally, we include the speaker text below the grid to provide further context to GPT-4V. The process is illustrated in Fig.~\ref{fig:video-caption}. %These frames are placed in a $3\times3$ grid in row-major order. Speaker text is placed below the grid to provide additional context to GPT-4V. The process is shown in the Fig.~\ref{fig:video-caption}. 

%Due to the rate limits of Vision API, all the utterances of the conversation cannot be passed in one request. We batch the conversation and get outputs independently from the Vision model. 
To accommodate the rate limits of the Vision API, we batch the utterances of a conversation and obtain outputs independently from the Vision model. We stitch all the outputs of a batched conversation into a single caption using GPT-3.5 (Appendix Fig.~\ref{fig:batched-caption-stitching}).

\subsubsection{Emotion Recognition}\label{approach2:emorec}
GPT tends to be uncontrollable when performing zero-shot recognition of emotions in conversations \cite{qin2023chatgpt} outputting emotions that are not a valid category of labels. To guide and control the process, we leverage in-context learning (ICL) by retrieving a conversation from the training set whose emotions are already annotated. The emotions in these conversations are explained by GPT-3.5 (Appendix Fig.~\ref{fig:emotion-explanation}). %The retrieved conversation and explanation serve as a demonstration for GPT to learn from to recognize emotions in conversation. Video caption is also provided as part of the prompt template (Appendix Fig.~\ref{fig:erc-context-prompt}). 
This retrieved conversation and its explanation serve as a demonstration for GPT to learn from, enabling it to recognize emotions in conversations more accurately. In addition, the prompt template includes the video caption as part of the input, as shown in Appendix Fig.~\ref{fig:erc-context-prompt}. 

To ensure effective ICL, it is important to provide general and descriptive examples that aid in solving the current task. %the provided examples should be general and descriptive to help solve the current task. 
In our approach, we sampled conversations from the training set containing all emotion categories. These conversations were stored as text-embedding-ada-002 embeddings \cite{neelakantan2022text} in a vector database. At test time, we compute the embedding for a conversation and retrieve the closest matching embedding from the database based on Euclidean distance. The retrieved embedding aids ICL in improving emotion understanding and recognition. %For a conversation at test time, we compute its embedding and find the closest embedding from the database based on Euclidean distance to be retrieved as an example.

\subsubsection{Cause Prediction}
Following the prediction of emotions, we predict the causes for each emotional utterance within a context window around that utterance. The bounds of the context window are given in Table~\ref{tab:context-bounds}. The bounds were informed by the distribution of the majority of relative positions of causes in the training set (Figure \ref{fig:rel-pos}).

\begin{table}[h!]
    \centering
    \small
    \begin{tabular}{ccc}
    \hline
        \textbf{Position} & \textbf{Previous} &  \textbf{Next}\\
    \hline
        Beginning & 0 & 2 \\
        End & 5 & 0 \\
        Middle & 5 & 2\\
    \hline
    \end{tabular}
    \caption{Context Window Bounds in each Direction}
    \label{tab:context-bounds}
\end{table}
 For predicting the causes of an utterance with emotion $e$ within a given context window $c$, we retrieve context windows containing utterances with the same emotion $e$ that exhibit semantic similarity to $c$. This retrieval is accomplished through the Euclidean distance comparison of text-embedding-ada-002 embeddings derived from the training data. The retrieved conversation's causes are explained by GPT-3.5 (Appendix Fig.~\ref{fig:cause-explanation}). Learning from the explained retrieved-context windows, cause prediction on $c$ can be performed by GPT-3.5. Video captions are also included in the prompt (Appendix Fig.~\ref{fig:cpred-context-prompt}), since the local window may have lost some broader context.

\subsection{Post-Processing}
\label{appendix:postproc}
In both our approaches, after getting the causes, we perform a post-processing step where we add the emotional utterance as its own cause which we call self-causes. This gives significant performance boosts as a majority of the causes are self-causes as pointed out in Appendix ~\ref{sec:appendixB}.

\section{Experimental setup}
\paragraph{Training details}
\label{section:training}
For approach 1, the data is split into train, test, and validation sets in the ratio 8:1:1. We use peft library \footnote{https://huggingface.co/docs/peft/en/index} for Parameter-Efficient Fine-Tuning.  Due to memory constraints, we fine-tune a 4-bit quantized Llama-2 model using bitsandbytes library \footnote{https://github.com/TimDettmers/bitsandbytes}. We report the details of the implementation for both approaches in Appendix ~\ref{sec:appendixC}. 
\paragraph{Evaluation metrics}
For evaluating, we report the precision, recall, F1-score, and weighted F1 which can be found on the competition website.\footnote{https://nustm.github.io/SemEval-2024\_ECAC/}

\section{Results and Discussion}
\paragraph{Main results}
Both of our approaches gave competitive rankings on the official leaderboard for subtask 2 as shown in Table \ref{tab:leaderboard}. In-context-learning GPT gave better results on the evaluation set compared to Fine-tuned Llama, thus our final position on the leaderboard was rank 4.    
\begin{table}[htb]
    \centering
    \small
    \begin{tabular}{lcc}
    \hline
        \textbf{System} & \textbf{w-avg F1} &  \textbf{F1}\\
    \hline
         1. Samsung Research China-Beijing & 0.3774 & 0.3870 \\
         2. NUS-Emo & 0.3460 & 0.3517 \\
         3. SZTU-MIPS & 0.3435 & 0.3434 \\
         \textbf{4. GPT-ICL} \textit{(Ours)} & \textbf{0.2758} & \textbf{0.2816} \\
         5. MotoMoto & 0.2584 & 0.2595 \\
         \textbf{6. Fine-tuned Llama} \textit{(Ours)} & \textbf{0.2558} & \textbf{0.2630}\\
    \hline
    \end{tabular}
    \caption{Leaderboard Results on Evaluation Data}
    \label{tab:leaderboard}
\end{table}

\paragraph{Ablation study}
\label{sec:ablation}
We conduct extensive ablation studies to measure the importance of the techniques we employ summarized in Table ~\ref{tab:val-set}. For these experiments, we use a subset of our test set containing 528 utterances. It can be seen that the performance of zero-shot Llama as well as GPT is the lowest. Instruction-tuning and ICL clearly improve the performance on the task, showcasing the significance of making LLMs context-aware when tackling downstream tasks. Adding self-causes improves performance in both zero-shot and context-aware cases highlighting their importance. The incorporation of video captions leads to poorer results in context-learning. The detailed table is in Appendix ~\ref{appendix:results}.
\begin{table}[htbp]
    \centering
    \small
    \begin{tabular}{lcc}
    \hline
        \textbf{Approach} &  \textbf{F1} &  \textbf{w-avg F1}\\
    \hline
        Zero-shot Llama &   &  \\
        \hspace{1em} - w/o self-causes &  0.117 &  0.116 \\
        \hspace{1em} - w/ self-causes &  0.222 & 0.215 \\
        \textbf{Instruction-tuned Llama} &   &\\
        \hspace{1em} - w/o self-causes &  0.325 &0.318 \\
        \hspace{1em} \textbf{- w/ self-causes} &  \textbf{0.364} &\textbf{0.352}\\
    \hline
        Zero-shot GPT &  &  \\
        \hspace{1em} - w/o self-causes &  0.100 &  0.097\\
        \hspace{1em} - w/ self-causes &  0.189 &  0.184\\
        \textbf{In-context-learning GPT} &   & \\
        \hspace{1em} - w/o self-causes w/o video &  0.286 & 0.296\\
        \hspace{1em} - w/o self-causes w/ video &  0.235 & 0.241\\
        \hspace{1em} \textbf{-w/ self-causes w/o video} &  \textbf{0.336} & \textbf{0.342}\\
        \hspace{1em} -w/ self-causes w/ video &  0.329 & 0.334\\
    \hline
    \end{tabular}
    \caption{Results on Validation Set.}
    \label{tab:val-set}
\end{table}
\vspace{-10pt}

\paragraph{Limitations}
Our approaches are specific to one dataset and may not generalize well to other datasets. Due to resource limitations, we fine-tune a Llama 13b parameter model instead of 70b and use QLoRA instead of updating all parameters. To save costs, we used GPT-3.5 model instead of GPT-4. Even with extensive prompt engineering, GPT models tend to hallucinate or give unstructured outputs, requiring retry repeatedly. 
% We are not able to predict self-causes accurately, hence the reliance on our post-processing step (Appendix \ref{appendix:postproc}).
\section{Conclusion}
We tackled the Multimodal ECAC task with a two-step framework of recognizing emotions first and then predicting their causes using LLMs. We implemented two approaches: a Llama-2 model which has been fine-tuned with instructions and a GPT model which solves the task by learning from demonstration examples in context. Conversation-level video captions were extracted to provide more context to LLMs. Our second approach was our best submission for the task, placing us at rank 4 with our first approach being placed at rank 6. Our results were under cost constraints and further investigation with larger Llama-2 models and GPT-4 with more sophisticated ICL approaches are a clear follow-up of our work.
% \section*{Acknowledgement}
%\bibliography{bibfile}

\begin{thebibliography}{60}
\expandafter\ifx\csname natexlab\endcsname\relax\def\natexlab#1{#1}\fi

\bibitem[{Abdullah et~al.(2021)Abdullah, Ameen, Sadeeq, and Zeebaree}]{abdullah2021multimodal}
Sharmeen M Saleem~Abdullah Abdullah, Siddeeq Y~Ameen Ameen, Mohammed~AM Sadeeq, and Subhi Zeebaree. 2021.
\newblock Multimodal emotion recognition using deep learning.
\newblock \emph{Journal of Applied Science and Technology Trends}, 2(02):52--58.

\bibitem[{Achiam et~al.(2023)Achiam, Adler, Agarwal, Ahmad, Akkaya, Aleman, Almeida, Altenschmidt, Altman, Anadkat et~al.}]{openai2023gpt4}
Josh Achiam, Steven Adler, Sandhini Agarwal, Lama Ahmad, Ilge Akkaya, Florencia~Leoni Aleman, Diogo Almeida, Janko Altenschmidt, Sam Altman, Shyamal Anadkat, et~al. 2023.
\newblock Gpt-4 technical report.
\newblock \emph{arXiv preprint arXiv:2303.08774}.

\bibitem[{Anil et~al.(2023)Anil, Dai, Firat, Johnson, Lepikhin, Passos, Shakeri, Taropa, Bailey, Chen et~al.}]{anil2023palm}
Rohan Anil, Andrew~M Dai, Orhan Firat, Melvin Johnson, Dmitry Lepikhin, Alexandre Passos, Siamak Shakeri, Emanuel Taropa, Paige Bailey, Zhifeng Chen, et~al. 2023.
\newblock Palm 2 technical report.
\newblock \emph{arXiv preprint arXiv:2305.10403}.

\bibitem[{Barros et~al.(2018)Barros, Churamani, Lakomkin, Siqueira, Sutherland, and Wermter}]{barros2018omg}
Pablo Barros, Nikhil Churamani, Egor Lakomkin, Henrique Siqueira, Alexander Sutherland, and Stefan Wermter. 2018.
\newblock The omg-emotion behavior dataset.
\newblock In \emph{2018 International Joint Conference on Neural Networks (IJCNN)}, pages 1--7. IEEE.

\bibitem[{Busso et~al.(2008)Busso, Bulut, Lee, Kazemzadeh, Mower, Kim, Chang, Lee, and Narayanan}]{busso2008iemocap}
Carlos Busso, Murtaza Bulut, Chi-Chun Lee, Abe Kazemzadeh, Emily Mower, Samuel Kim, Jeannette~N Chang, Sungbok Lee, and Shrikanth~S Narayanan. 2008.
\newblock Iemocap: Interactive emotional dyadic motion capture database.
\newblock \emph{Language resources and evaluation}, 42:335--359.

\bibitem[{Chen et~al.(2020)Chen, Hou, Li, Wu, and Zhang}]{chen2020end}
Ying Chen, Wenjun Hou, Shoushan Li, Caicong Wu, and Xiaoqiang Zhang. 2020.
\newblock End-to-end emotion-cause pair extraction with graph convolutional network.
\newblock In \emph{Proceedings of the 28th International Conference on Computational Linguistics}, pages 198--207.

\bibitem[{Chen et~al.(2010)Chen, Lee, Li, and Huang}]{chen2010emotion}
Ying Chen, Sophia Yat~Mei Lee, Shoushan Li, and Chu-Ren Huang. 2010.
\newblock Emotion cause detection with linguistic constructions.
\newblock In \emph{Proceedings of the 23rd International Conference on Computational Linguistics (Coling 2010)}, pages 179--187.

\bibitem[{Chou et~al.(2017)Chou, Lin, Chang, Li, Ma, and Lee}]{chou2017nnime}
Huang-Cheng Chou, Wei-Cheng Lin, Lien-Chiang Chang, Chyi-Chang Li, Hsi-Pin Ma, and Chi-Chun Lee. 2017.
\newblock Nnime: The nthu-ntua chinese interactive multimodal emotion corpus.
\newblock In \emph{2017 Seventh International Conference on Affective Computing and Intelligent Interaction (ACII)}, pages 292--298. IEEE.

\bibitem[{Dettmers et~al.(2024)Dettmers, Pagnoni, Holtzman, and Zettlemoyer}]{dettmers2024qlora}
Tim Dettmers, Artidoro Pagnoni, Ari Holtzman, and Luke Zettlemoyer. 2024.
\newblock Qlora: Efficient finetuning of quantized llms.
\newblock \emph{Advances in Neural Information Processing Systems}, 36.

\bibitem[{Dong et~al.(2023)Dong, Li, Dai, Zheng, Wu, Chang, Sun, Xu, Li, and Sui}]{Dong2023ASF}
Qingxiu Dong, Lei Li, Damai Dai, Ce~Zheng, Zhiyong Wu, Baobao Chang, Xu~Sun, Jingjing Xu, Lei Li, and Zhifang Sui. 2023.
\newblock \href {https://api.semanticscholar.org/CorpusID:263886074} {A survey for in-context learning}.
\newblock \emph{ArXiv}, abs/2301.00234.

\bibitem[{Douze et~al.(2024)Douze, Guzhva, Deng, Johnson, Szilvasy, Mazaré, Lomeli, Hosseini, and Jégou}]{douze2024faiss}
Matthijs Douze, Alexandr Guzhva, Chengqi Deng, Jeff Johnson, Gergely Szilvasy, Pierre-Emmanuel Mazaré, Maria Lomeli, Lucas Hosseini, and Hervé Jégou. 2024.
\newblock \href {http://arxiv.org/abs/2401.08281} {The faiss library}.

\bibitem[{Fan et~al.(2020)Fan, Yuan, Du, Gui, Yang, and Xu}]{fan2020transition}
Chuang Fan, Chaofa Yuan, Jiachen Du, Lin Gui, Min Yang, and Ruifeng Xu. 2020.
\newblock Transition-based directed graph construction for emotion-cause pair extraction.
\newblock In \emph{Proceedings of the 58th Annual Meeting of the Association for Computational Linguistics}, pages 3707--3717.

\bibitem[{Firdaus et~al.(2020)Firdaus, Chauhan, Ekbal, and Bhattacharyya}]{firdaus2020meisd}
Mauajama Firdaus, Hardik Chauhan, Asif Ekbal, and Pushpak Bhattacharyya. 2020.
\newblock Meisd: A multimodal multi-label emotion, intensity and sentiment dialogue dataset for emotion recognition and sentiment analysis in conversations.
\newblock In \emph{Proceedings of the 28th international conference on computational linguistics}, pages 4441--4453.

\bibitem[{Fu et~al.(2023)Fu, Yuan, Zhang, and Cao}]{fu2023emotion}
Yao Fu, Shaoyang Yuan, Chi Zhang, and Juan Cao. 2023.
\newblock Emotion recognition in conversations: A survey focusing on context, speaker dependencies, and fusion methods.
\newblock \emph{Electronics}, 12(22):4714.

\bibitem[{Ghazi et~al.(2015)Ghazi, Inkpen, and Szpakowicz}]{ghazi2015detecting}
Diman Ghazi, Diana Inkpen, and Stan Szpakowicz. 2015.
\newblock Detecting emotion stimuli in emotion-bearing sentences.
\newblock In \emph{Computational Linguistics and Intelligent Text Processing: 16th International Conference, CICLing 2015, Cairo, Egypt, April 14-20, 2015, Proceedings, Part II 16}, pages 152--165. Springer.

\bibitem[{Gui et~al.(2018)Gui, Xu, Wu, Lu, and Zhou}]{gui2018event}
Lin Gui, Ruifeng Xu, Dongyin Wu, Qin Lu, and Yu~Zhou. 2018.
\newblock Event-driven emotion cause extraction with corpus construction.
\newblock In \emph{Social Media Content Analysis: Natural Language Processing and Beyond}, pages 145--160. World Scientific.

\bibitem[{Hsu et~al.(2018)Hsu, Chen, Kuo, Huang, and Ku}]{hsu2018emotionlines}
Chao-Chun Hsu, Sheng-Yeh Chen, Chuan-Chun Kuo, Ting-Hao Huang, and Lun-Wei Ku. 2018.
\newblock Emotionlines: An emotion corpus of multi-party conversations.
\newblock In \emph{Proceedings of the Eleventh International Conference on Language Resources and Evaluation (LREC 2018)}.

\bibitem[{Hu et~al.(2021)Hu, Lu, and Zhao}]{hu2021fss}
Guimin Hu, Guangming Lu, and Yi~Zhao. 2021.
\newblock Fss-gcn: A graph convolutional networks with fusion of semantic and structure for emotion cause analysis.
\newblock \emph{Knowledge-Based Systems}, 212:106584.

\bibitem[{Imran et~al.(2023)Imran, Chatterjee, and Damevski}]{imran2023uncovering}
Mia~Mohammad Imran, Preetha Chatterjee, and Kostadin Damevski. 2023.
\newblock Uncovering the causes of emotions in software developer communication using zero-shot llms.
\newblock \emph{arXiv preprint arXiv:2312.09731}.

\bibitem[{Lei et~al.(2023)Lei, Dong, Wang, Wang, and Wang}]{lei2023instructerc}
Shanglin Lei, Guanting Dong, Xiaoping Wang, Keheng Wang, and Sirui Wang. 2023.
\newblock \href {http://arxiv.org/abs/2309.11911} {Instructerc: Reforming emotion recognition in conversation with a retrieval multi-task llms framework}.

\bibitem[{Li and Xu(2014)}]{li2014text}
Weiyuan Li and Hua Xu. 2014.
\newblock Text-based emotion classification using emotion cause extraction.
\newblock \emph{Expert Systems with Applications}, 41(4):1742--1749.

\bibitem[{Li et~al.(2021)Li, Gao, Feng, Zhang, and Wang}]{li2021boundary}
Xiangju Li, Wei Gao, Shi Feng, Yifei Zhang, and Daling Wang. 2021.
\newblock Boundary detection with bert for span-level emotion cause analysis.
\newblock In \emph{Findings of the Association for Computational Linguistics: ACL-IJCNLP 2021}, pages 676--682.

\bibitem[{Li et~al.(2023)Li, Wang, and Cui}]{li2023decoupled}
Yong Li, Yuanzhi Wang, and Zhen Cui. 2023.
\newblock Decoupled multimodal distilling for emotion recognition.
\newblock In \emph{Proceedings of the IEEE/CVF Conference on Computer Vision and Pattern Recognition}, pages 6631--6640.

\bibitem[{Lin et~al.(2023)Lin, Ahmed, Li, Lin, Azarnasab, Yang, Wang, Liang, Liu, Lu, Liu, and Wang}]{Lin2023MMVIDAV}
Kevin Lin, Faisal Ahmed, Linjie Li, Chung-Ching Lin, Ehsan Azarnasab, Zhengyuan Yang, Jianfeng Wang, Lin Liang, Zicheng Liu, Yumao Lu, Ce~Liu, and Lijuan Wang. 2023.
\newblock \href {https://api.semanticscholar.org/CorpusID:264806489} {Mm-vid: Advancing video understanding with gpt-4v(ision)}.
\newblock \emph{ArXiv}, abs/2310.19773.

\bibitem[{Mittal et~al.(2021)Mittal, Mathur, Bera, and Manocha}]{mittal2021affect2mm}
Trisha Mittal, Puneet Mathur, Aniket Bera, and Dinesh Manocha. 2021.
\newblock Affect2mm: Affective analysis of multimedia content using emotion causality.
\newblock In \emph{Proceedings of the IEEE/CVF Conference on Computer Vision and Pattern Recognition}, pages 5661--5671.

\bibitem[{Neelakantan et~al.(2022)Neelakantan, Xu, Puri, Radford, Han, Tworek, Yuan, Tezak, Kim, Hallacy et~al.}]{neelakantan2022text}
Arvind Neelakantan, Tao Xu, Raul Puri, Alec Radford, Jesse~Michael Han, Jerry Tworek, Qiming Yuan, Nikolas Tezak, Jong~Wook Kim, Chris Hallacy, et~al. 2022.
\newblock Text and code embeddings by contrastive pre-training.
\newblock \emph{arXiv preprint arXiv:2201.10005}.

\bibitem[{Peng et~al.(2022)Peng, Cao, Zhou, Ouyang, Yang, Li, Jia, and Yu}]{peng2022survey}
Sancheng Peng, Lihong Cao, Yongmei Zhou, Zhouhao Ouyang, Aimin Yang, Xinguang Li, Weijia Jia, and Shui Yu. 2022.
\newblock A survey on deep learning for textual emotion analysis in social networks.
\newblock \emph{Digital Communications and Networks}, 8(5):745--762.

\bibitem[{Pereira et~al.(2022)Pereira, Moniz, and Carvalho}]{pereira2022deep}
Patr{\'\i}cia Pereira, Helena Moniz, and Joao~Paulo Carvalho. 2022.
\newblock Deep emotion recognition in textual conversations: A survey.
\newblock \emph{arXiv preprint arXiv:2211.09172}.

\bibitem[{Poria et~al.(2019)Poria, Hazarika, Majumder, Naik, Cambria, and Mihalcea}]{poria2019meld}
Soujanya Poria, Devamanyu Hazarika, Navonil Majumder, Gautam Naik, Erik Cambria, and Rada Mihalcea. 2019.
\newblock Meld: A multimodal multi-party dataset for emotion recognition in conversations.
\newblock In \emph{Proceedings of the 57th Annual Meeting of the Association for Computational Linguistics}, pages 527--536.

\bibitem[{Qin et~al.(2023)Qin, Zhang, Zhang, Chen, Yasunaga, and Yang}]{qin2023chatgpt}
Chengwei Qin, Aston Zhang, Zhuosheng Zhang, Jiaao Chen, Michihiro Yasunaga, and Diyi Yang. 2023.
\newblock Is chatgpt a general-purpose natural language processing task solver?
\newblock \emph{arXiv preprint arXiv:2302.06476}.

\bibitem[{Sebe et~al.(2005)Sebe, Cohen, Gevers, and Huang}]{sebe2005multimodal}
Nicu Sebe, Ira Cohen, Theo Gevers, and Thomas~S Huang. 2005.
\newblock Multimodal approaches for emotion recognition: a survey.
\newblock In \emph{Internet Imaging VI}, volume 5670, pages 56--67. SPIE.

\bibitem[{Shoumy et~al.(2020)Shoumy, Ang, Seng, Rahaman, and Zia}]{shoumy2020multimodal}
Nusrat~J Shoumy, Li-Minn Ang, Kah~Phooi Seng, DM~Motiur Rahaman, and Tanveer Zia. 2020.
\newblock Multimodal big data affective analytics: A comprehensive survey using text, audio, visual and physiological signals.
\newblock \emph{Journal of Network and Computer Applications}, 149:102447.

\bibitem[{Singh et~al.(2021)Singh, Hingane, Wani, and Modi}]{singh2021end}
Aaditya Singh, Shreeshail Hingane, Saim Wani, and Ashutosh Modi. 2021.
\newblock An end-to-end network for emotion-cause pair extraction.
\newblock In \emph{Proceedings of the Eleventh Workshop on Computational Approaches to Subjectivity, Sentiment and Social Media Analysis}, pages 84--91.

\bibitem[{Touvron et~al.(2023)Touvron, Lavril, Izacard, Martinet, Lachaux, Lacroix, Rozi{\`e}re, Goyal, Hambro, Azhar et~al.}]{touvron2023llama}
Hugo Touvron, Thibaut Lavril, Gautier Izacard, Xavier Martinet, Marie-Anne Lachaux, Timoth{\'e}e Lacroix, Baptiste Rozi{\`e}re, Naman Goyal, Eric Hambro, Faisal Azhar, et~al. 2023.
\newblock Llama: Open and efficient foundation language models.
\newblock \emph{arXiv preprint arXiv:2302.13971}.

\bibitem[{Wang et~al.(2021)Wang, Ding, Xia, Li, and Yu}]{wang2021multimodal}
Fanfan Wang, Zixiang Ding, Rui Xia, Zhaoyu Li, and Jianfei Yu. 2021.
\newblock \href {http://arxiv.org/abs/2110.08020} {Multimodal emotion-cause pair extraction in conversations}.

\bibitem[{Wang et~al.(2023{\natexlab{a}})Wang, Ding, Xia, Li, and Yu}]{DBLP:journals/taffco/WangDXLY23}
Fanfan Wang, Zixiang Ding, Rui Xia, Zhaoyu Li, and Jianfei Yu. 2023{\natexlab{a}}.
\newblock \href {https://doi.org/10.1109/TAFFC.2022.3226559} {Multimodal emotion-cause pair extraction in conversations}.
\newblock \emph{{IEEE} Trans. Affect. Comput.}, 14(3):1832--1844.

\bibitem[{Wang et~al.(2024{\natexlab{a}})Wang, Ma, Yu, Xia, and Cambria}]{ECAC2024SemEval}
Fanfan Wang, Heqing Ma, Jianfei Yu, Rui Xia, and Erik Cambria. 2024{\natexlab{a}}.
\newblock Semeval-2024 task 3: Multimodal emotion cause analysis in conversations.
\newblock In \emph{Proceedings of the 18th International Workshop on Semantic Evaluation (SemEval-2024)}.

\bibitem[{Wang et~al.(2024{\natexlab{b}})Wang, Chen, Chen, Wu, Zhu, Zeng, Luo, Lu, Zhou, Qiao et~al.}]{wang2024visionllm}
Wenhai Wang, Zhe Chen, Xiaokang Chen, Jiannan Wu, Xizhou Zhu, Gang Zeng, Ping Luo, Tong Lu, Jie Zhou, Yu~Qiao, et~al. 2024{\natexlab{b}}.
\newblock Visionllm: Large language model is also an open-ended decoder for vision-centric tasks.
\newblock \emph{Advances in Neural Information Processing Systems}, 36.

\bibitem[{Wang et~al.(2023{\natexlab{b}})Wang, Ma, and Chen}]{wang2023augmenting}
Yubo Wang, Xueguang Ma, and Wenhu Chen. 2023{\natexlab{b}}.
\newblock \href {http://arxiv.org/abs/2309.02233} {Augmenting black-box llms with medical textbooks for clinical question answering}.

\bibitem[{Wang et~al.(2023{\natexlab{c}})Wang, Li, Yu, and Hu}]{wang2023knowledge}
Yuwei Wang, Yuling Li, Kui Yu, and Yimin Hu. 2023{\natexlab{c}}.
\newblock Knowledge-enhanced hierarchical transformers for emotion-cause pair extraction.
\newblock In \emph{Pacific-Asia Conference on Knowledge Discovery and Data Mining}, pages 112--123. Springer.

\bibitem[{Wang et~al.(2023{\natexlab{d}})Wang, Xie, Ding, Feng, and Xia}]{wang2023chatgpt}
Zengzhi Wang, Qiming Xie, Zixiang Ding, Yi~Feng, and Rui Xia. 2023{\natexlab{d}}.
\newblock Is chatgpt a good sentiment analyzer? a preliminary study.
\newblock \emph{arXiv preprint arXiv:2304.04339}.

\bibitem[{Wei et~al.(2022{\natexlab{a}})Wei, Tay, Bommasani, Raffel, Zoph, Borgeaud, Yogatama, Bosma, Zhou, Metzler et~al.}]{wei2022emergent}
Jason Wei, Yi~Tay, Rishi Bommasani, Colin Raffel, Barret Zoph, Sebastian Borgeaud, Dani Yogatama, Maarten Bosma, Denny Zhou, Donald Metzler, et~al. 2022{\natexlab{a}}.
\newblock Emergent abilities of large language models.
\newblock \emph{arXiv preprint arXiv:2206.07682}.

\bibitem[{Wei et~al.(2022{\natexlab{b}})Wei, Wang, Schuurmans, Bosma, Xia, Chi, Le, Zhou et~al.}]{wei2022chain}
Jason Wei, Xuezhi Wang, Dale Schuurmans, Maarten Bosma, Fei Xia, Ed~Chi, Quoc~V Le, Denny Zhou, et~al. 2022{\natexlab{b}}.
\newblock Chain-of-thought prompting elicits reasoning in large language models.
\newblock \emph{Advances in Neural Information Processing Systems}, 35:24824--24837.

\bibitem[{Wei et~al.(2020)Wei, Zhao, and Mao}]{wei2020effective}
Penghui Wei, Jiahao Zhao, and Wenji Mao. 2020.
\newblock Effective inter-clause modeling for end-to-end emotion-cause pair extraction.
\newblock In \emph{Proceedings of the 58th Annual Meeting of the Association for Computational Linguistics}, pages 3171--3181.

\bibitem[{W{\"o}llmer et~al.(2013)W{\"o}llmer, Weninger, Knaup, Schuller, Sun, Sagae, and Morency}]{wollmer2013youtube}
Martin W{\"o}llmer, Felix Weninger, Tobias Knaup, Bj{\"o}rn Schuller, Congkai Sun, Kenji Sagae, and Louis-Philippe Morency. 2013.
\newblock Youtube movie reviews: Sentiment analysis in an audio-visual context.
\newblock \emph{IEEE Intelligent Systems}, 28(3):46--53.

\bibitem[{Wu et~al.(2024)Wu, Shen, Zhang, and Cai}]{wu2024enhancing}
Jialiang Wu, Yi~Shen, Ziheng Zhang, and Longjun Cai. 2024.
\newblock Enhancing large language model with decomposed reasoning for emotion cause pair extraction.
\newblock \emph{arXiv preprint arXiv:2401.17716}.

\bibitem[{Xia and Ding(2019)}]{xia2019emotion}
Rui Xia and Zixiang Ding. 2019.
\newblock Emotion-cause pair extraction: A new task to emotion analysis in texts.
\newblock In \emph{Proceedings of the 57th Annual Meeting of the Association for Computational Linguistics}, pages 1003--1012.

\bibitem[{Yada et~al.(2017)Yada, Ikeda, Hoashi, and Kageura}]{yada2017bootstrap}
Shuntaro Yada, Kazushi Ikeda, Keiichiro Hoashi, and Kyo Kageura. 2017.
\newblock A bootstrap method for automatic rule acquisition on emotion cause extraction.
\newblock In \emph{2017 IEEE International Conference on Data Mining Workshops (ICDMW)}, pages 414--421. IEEE.

\bibitem[{Yang et~al.(2024)Yang, Zhang, Li, Marta, Batool, and Folkesson}]{Yang_2024_WACV}
Yi~Yang, Qingwen Zhang, Ci~Li, Daniel Sim\~oes Marta, Nazre Batool, and John Folkesson. 2024.
\newblock Human-centric autonomous systems with llms for user command reasoning.
\newblock In \emph{Proceedings of the IEEE/CVF Winter Conference on Applications of Computer Vision (WACV) Workshops}, pages 988--994.

\bibitem[{Yang et~al.(2023)Yang, Li, Lin, Wang, Lin, Liu, and Wang}]{Yang2023lmms}
Zhengyuan Yang, Linjie Li, Kevin Lin, Jianfeng Wang, Chung-Ching Lin, Zicheng Liu, and Lijuan Wang. 2023.
\newblock \href {https://api.semanticscholar.org/CorpusID:263310951} {The dawn of lmms: Preliminary explorations with gpt-4v(ision)}.
\newblock \emph{ArXiv}, abs/2309.17421.

\bibitem[{Yu et~al.(2020)Yu, Xu, Meng, Zhu, Ma, Wu, Zou, and Yang}]{yu2020ch}
Wenmeng Yu, Hua Xu, Fanyang Meng, Yilin Zhu, Yixiao Ma, Jiele Wu, Jiyun Zou, and Kaicheng Yang. 2020.
\newblock Ch-sims: A chinese multimodal sentiment analysis dataset with fine-grained annotation of modality.
\newblock In \emph{Proceedings of the 58th annual meeting of the association for computational linguistics}, pages 3718--3727.

\bibitem[{Zadeh et~al.(2016)Zadeh, Zellers, Pincus, and Morency}]{zadeh2016mosi}
Amir Zadeh, Rowan Zellers, Eli Pincus, and Louis-Philippe Morency. 2016.
\newblock Mosi: multimodal corpus of sentiment intensity and subjectivity analysis in online opinion videos.
\newblock \emph{arXiv preprint arXiv:1606.06259}.

\bibitem[{Zhang et~al.(2023)Zhang, Yang, Chen, Zhang, Leng, and Zhao}]{zhang2023deep}
Shiqing Zhang, Yijiao Yang, Chen Chen, Xingnan Zhang, Qingming Leng, and Xiaoming Zhao. 2023.
\newblock Deep learning-based multimodal emotion recognition from audio, visual, and text modalities: A systematic review of recent advancements and future prospects.
\newblock \emph{Expert Systems with Applications}, page 121692.

\bibitem[{Zhang and Li(2023)}]{zhang2023cross}
Xiaoheng Zhang and Yang Li. 2023.
\newblock A cross-modality context fusion and semantic refinement network for emotion recognition in conversation.
\newblock In \emph{Proceedings of the 61st Annual Meeting of the Association for Computational Linguistics (Volume 1: Long Papers)}, pages 13099--13110.

\bibitem[{Zhang et~al.(2024)Zhang, Wang, Wu, Tiwari, Li, Wang, and Qin}]{zhang2024dialoguellm}
Yazhou Zhang, Mengyao Wang, Youxi Wu, Prayag Tiwari, Qiuchi Li, Benyou Wang, and Jing Qin. 2024.
\newblock \href {http://arxiv.org/abs/2310.11374} {Dialoguellm: Context and emotion knowledge-tuned large language models for emotion recognition in conversations}.

\bibitem[{Zhao et~al.(2023)Zhao, Zhao, Li, and Qin}]{zhao2023knowledge}
Weixiang Zhao, Yanyan Zhao, Zhuojun Li, and Bing Qin. 2023.
\newblock Knowledge-bridged causal interaction network for causal emotion entailment.
\newblock In \emph{Proceedings of the AAAI Conference on Artificial Intelligence}, volume~37, pages 14020--14028.

\bibitem[{Zheng et~al.(2023{\natexlab{a}})Zheng, Ji, Li, Fei, Wu, Li, Li, and Teng}]{zheng2023ecqed}
Li~Zheng, Donghong Ji, Fei Li, Hao Fei, Shengqiong Wu, Jingye Li, Bobo Li, and Chong Teng. 2023{\natexlab{a}}.
\newblock Ecqed: Emotion-cause quadruple extraction in dialogs.
\newblock \emph{arXiv preprint arXiv:2306.03969}.

\bibitem[{Zheng et~al.(2023{\natexlab{b}})Zheng, Yu, Xia, and Wang}]{zheng2023facial}
Wenjie Zheng, Jianfei Yu, Rui Xia, and Shijin Wang. 2023{\natexlab{b}}.
\newblock A facial expression-aware multimodal multi-task learning framework for emotion recognition in multi-party conversations.
\newblock In \emph{Proceedings of the 61st Annual Meeting of the Association for Computational Linguistics (Volume 1: Long Papers)}, pages 15445--15459.

\bibitem[{Zheng et~al.(2022)Zheng, Liu, Zhang, Wang, and Wang}]{zheng2022ueca}
Xiaopeng Zheng, Zhiyue Liu, Zizhen Zhang, Zhaoyang Wang, and Jiahai Wang. 2022.
\newblock Ueca-prompt: Universal prompt for emotion cause analysis.
\newblock In \emph{Proceedings of the 29th International Conference on Computational Linguistics}, pages 7031--7041.

\bibitem[{Zhong et~al.(2019)Zhong, Wang, and Miao}]{zhong2019knowledge}
Peixiang Zhong, Di~Wang, and Chunyan Miao. 2019.
\newblock Knowledge-enriched transformer for emotion detection in textual conversations.
\newblock In \emph{Proceedings of the 2019 Conference on Empirical Methods in Natural Language Processing and the 9th International Joint Conference on Natural Language Processing (EMNLP-IJCNLP)}, pages 165--176.

\end{thebibliography}
%\bibliographystyle{acl_natbib}

\appendix

\section{Related Work}
\label{appendix:related_work}
 Our system is designed to prioritize Subtask 2 which is directly related to text-based and multimodal ECA. In the following sections, we will present relevant research that addresses both unimodal (text-based) and multimodal ECA.

 \subsection*{Text-based ECA}Advancements in text-based ECA~\citep{xia2019emotion,hsu2018emotionlines,peng2022survey,pereira2022deep} have made significant strides within the field of sentiment analysis. The task on emotion cause extraction (ECE) was initially proposed by \citet{chen2010emotion} on a Chinese corpus. Several studies~\citep{,li2014text,ghazi2015detecting,yada2017bootstrap} have explored ECE task, using both rule-based and machine learning approaches that operate at the phrase or word level of the text data. Furthermore, \citep{gui2018event} reformulated the ECE task as a clause-level classification problem and constructed a Chinese emotion-cause corpus based on the news data. Considering the effectiveness of clause-level units in indicating emotions, \citet{xia2019emotion} introduced the task of Emotion-Cause Pair Extraction (ECPE) for extracting potential emotion-cause pairs from texts. Numerous deep learning models~\citep{zhong2019knowledge,wei2020effective,chen2020end,singh2021end,li2021boundary,wang2023knowledge} have been developed to address ECPE tasks. Additionally, graph-based approaches~\citep{zheng2023ecqed,hu2021fss,zhao2023knowledge} that utilize graphs to model dialog context and capture interactions between speakers and utterances hold significant potential. The focus on transformer models and the rapid progress in LLMs such as ChatGPT\footnote{https://chat.openai.com/} and Llama~\citep{touvron2023llama}, have significantly boosted the performance of various NLP tasks~\citep{imran2023uncovering} including ECPE~\citep{wang2023chatgpt,imran2023uncovering,wu2024enhancing,zheng2022ueca}.% We adopt an idea of using generative AI models for.....
 \subsection*{Multimodal ECA}
Given the strong association between facial cues and emotion, integrating modalities to improve emotion recognition has attracted a lot of attention~\citep{sebe2005multimodal,li2023decoupled,zhang2023deep,fu2023emotion}. Several key multimodal datasets~\citep{wollmer2013youtube,zadeh2016mosi,chou2017nnime,barros2018omg,poria2019meld,yu2020ch} have emerged to support and advance research. The availability of open conversation data has facilitated the expansion of multimodal conversation datasets, which includes various types of conversations such as dyadic interactions~\citep{busso2008iemocap}, and multi-participant communications~\citep{hsu2018emotionlines,poria2019meld,firdaus2020meisd,zheng2023facial}.

\subsection*{Large Language Models}
The emergence of Large Language Models such as GPT-4 \cite{openai2023gpt4}, Llama \cite{touvron2023llama}, PaLM \cite{anil2023palm}, etc. has transformed the research landscape. Recently, there has been a surge in the application of LLMs to a multitude of domains. \citet{zhang2024dialoguellm} extend their capabilities to the task of emotion recognition where they fine-tune a Llama 2-7 billion parameter model for emotion prediction. \citet{lei2023instructerc} introduce a retrieval template module along with speaker identification and emotion-impact prediction tasks to improve the performance of LLM. In our work, as part of approach 1, we develop two distinct LLM-based experts separately for emotion and cause prediction. 

\citet{qin2023chatgpt} investigated the task of zero-shot emotion cause prediction using ChatGPT with limited success. Recently, a new paradigm of in-context learning (ICL) \cite{Dong2023ASF} has emerged for LLMs that involves learning from a few examples to solve a variety of complex reasoning tasks \cite{wei2022chain}, \cite{wei2022emergent}. \citet{wu2024enhancing} proposed a Chain of Thought (CoT) \cite{wei2022chain} approach for emotion cause pair extraction. Our approach 2 extends the idea of ICL towards solving the task of multimodal emotion cause pair extraction in two steps.
  
\section{Implementation details}
\label{sec:appendixC}
\subsection{Training details for Llama}
Both emotion and cause prediction training used one Nvidia A100 40GB GPU for training (Available on \href{https://colab.research.google.com/signup}{Google Colab Pro} priced at \$11.8/month). We train for one epoch due to constraints on Colab usage with gradient accumulation steps as 8 with an effective batch size of 8. A cosine learning rate scheduler and Adam optimizer are used. Inference is performed using two Tesla T4 16GB GPUs (Available on \href{https://www.kaggle.com/}{Kaggle} for free (30 hrs/month)).

The long context length of 4096 tokens of the Llama 2 models, allows us to include the entire conversation as context and input that to the model. We perform experiments to analyze the maximum token counts in the dataset and observe that they do not exceed 1600 as shown in Figure \ref{fig:token-counts}. In case the token count exceeds the limit for the LLM we can use a window of utterances around the given utterance as context for predicting its emotion.
\begin{table}[htb]
    \centering
    \small{
    \begin{tabular}{cc}
    \hline
       \textbf{Hyperparameter}  & \textbf{Value} \\
       \hline
        Lora alpha & 16 \\
        Lora dropout & 0.1 \\
        Attention heads & 16 \\
        Learning rate & 1e-3 \\
        Epochs & 1 \\
        LR scheduler & cosine \\
        Warmup ratio & 0.03 \\
        Weight decay & 0.001 \\
        \hline
    \end{tabular}
    }
    \caption{Hyperparameters for fine-tuning}
    \label{tab:hyperparameters}
\end{table}

\begin{figure}
    \centering
    \includegraphics[width=1.1\linewidth]{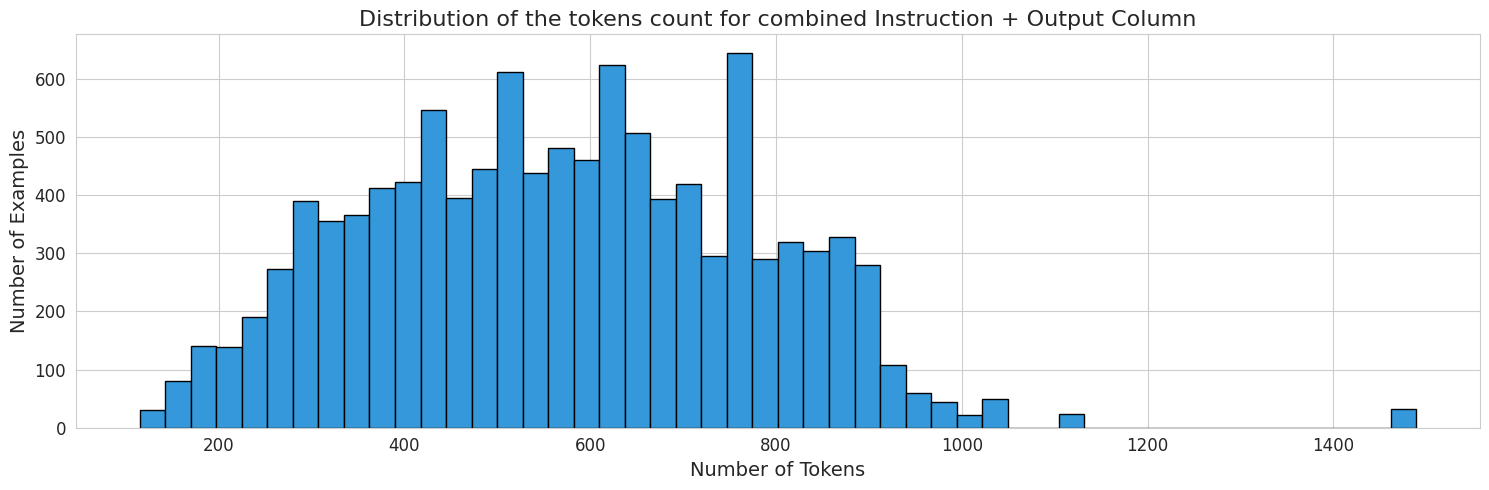}
    \caption{Distribution of token counts for Llama tokenizer}
    \label{fig:token-counts}
\end{figure}

\subsection{Details for in-context learning GPT}
\label{sec:appendixC.2}
We use the LangChain\footnote{https://github.com/langchain-ai/langchain} library to implement our three pipelines: video captioning, emotion recognition, and cause prediction. We use the interface provided by LangChain to communicate with OpenAI's API models detailed in Table \ref{tab:openai-models}. 

\begin{table}[htb]
    \centering
    \small{
    \begin{tabular}{cc}
    \hline
       \textbf{Model}  & \textbf{API Name}\\
        \hline
            GPT-4V & gpt-4-vision-preview \\
            GPT-3.5 & gpt-3.5-turbo-1106 \\
            Embeddings & text-embedding-ada-002 \\
        \hline
    \end{tabular}
    }
    \caption{OpenAI API Model Names}
    \label{tab:openai-models}
\end{table}

\paragraph{Vector databases} For creating vector databases, we use the FAISS Library \cite{douze2024faiss}. We created a FAISS index containing embeddings of 12 conversations from the training set which contains all emotion categories. For cause prediction, we created a FAISS index for each of the 6 emotion categories and 3 possible positions of emotional utterance giving us a total of 18 indices. Each of these indices contained embeddings of context windows (bounds defined in Table \ref{tab:context-bounds}) from the training set corresponding to each emotion and position. 

\begin{table*}[t!]
    \centering
    \small
    \begin{tabular}{lcccccc}
    \hline
        \textbf{Approach} & \textbf{P} & \textbf{R} & \textbf{F1} & \textbf{w-P} & \textbf{w-R} & \textbf{w-avg F1}\\
    \hline
        Zero-shot Llama w/o self-causes & 0.089 & 0.168 & 0.117 & 0.090 & 0.168 & 0.116 \\
        Zero-shot Llama w/ self-causes & 0.157 & 0.372 & 0.222 & 0.152 &0.372 &0.215 \\
        Instruction-tuned Llama w/o self-causes & 0.351 & 0.304 & 0.325 &0.335 &0.304 &0.318 \\
        \textbf{Instruction-tuned Llama w/ self-causes} &  \textbf{0.360} &\textbf{0.367} &\textbf{0.364} &\textbf{0.342} &\textbf{0.367} &\textbf{0.352}\\
    \hline
        Zero-shot GPT w/o self-causes & 0.081 & 0.130 & 0.100 & 0.087 & 0.130 & 0.097\\
        Zero-shot GPT w/ self-causes & 0.140 & 0.290 & 0.189 & 0.149 & 0.290 & 0.184\\
        In-context-learning GPT w/o video captions w/o self-causes & 0.259 & 0.319 & 0.286 & 0.283 & 0.319 & 0.296\\
       \textbf{In-context-learning GPT w/o video captions w/ self-cause}s & \textbf{0.270} & \textbf{0.445 }& \textbf{0.336} & \textbf{0.287} & \textbf{0.445} & \textbf{0.342}\\ 
        In-context-learning GPT w/o self-causes & 0.216 & 0.256 & 0.235 & 0.241 & 0.256 & 0.241\\
        In-context-learning GPT w/ self-causes & 0.261 & 0.445 & 0.329 & 0.280 & 0.445 & 0.334\\
    \hline
    \end{tabular}
    \caption{Results on Validation Set. P: precision, R: recall, w: weighted.}
    \label{tab:val-set-full}
\end{table*}

\section{Detailed results}
\label{appendix:results}
The detailed results on precision, recall, and F1-scores are given in Table ~\ref{tab:val-set-full}.

\section{Error Analysis}
\label{sec:error_analysis}
We conduct error analysis for the output of emotion recognition using the two approaches. The performance of zero-shot Llama is extremely poor where the model predicts the label joy for almost all utterances (Fig.~\ref{fig:cm1}). On adding the conversational context, the model can identify the emotional nuances better, yet often predicts joy or surprise for neutral (Fig.~\ref{fig:cm2}). Instruction fine-tuning significantly boosts performance where the model can now differentiate distinct emotions (Fig.~\ref{fig:cm3}). The performance on disgust and fear is low due to the class-imbalance problem. In our test subset, the support of disgust and fear is only 13, as shown in Table ~\ref{tab:emotion_results}. We observed similar trends in the case of our second approach. Zero-shot GPT (Fig.~\ref{fig:zero-shot-cm}) tends to only identify the neutral utterances accurately and fails in other categories. The incorporation of in-context learning (Fig.~\ref{fig:gpt-icl-cm}) improves the accuracy in identifying different emotion categories but there is little to no improvement in identifying disgust or anger utterances.

\begin{figure}[h]
    \centering
    \includegraphics[width=\linewidth]{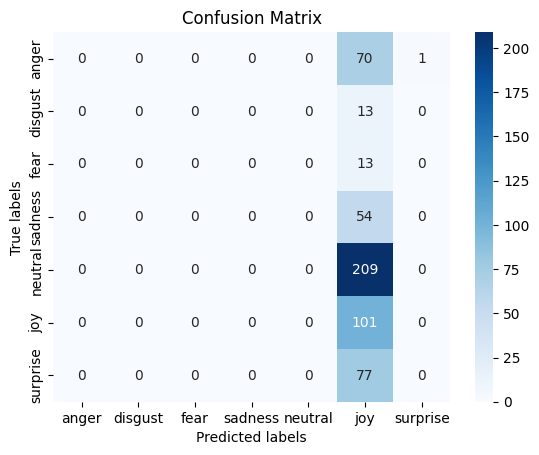}
    \caption{Confusion matrix for zero-shot emotion recognition without context using Llama}
    \label{fig:cm1}
\end{figure}
\begin{figure}[h]
    \centering
    \includegraphics[width=\linewidth]{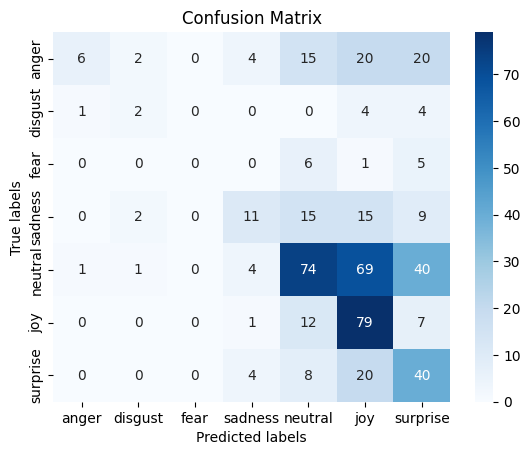}
    \caption{Confusion matrix for zero-shot emotion recognition with context using Llama}
    \label{fig:cm2}
\end{figure}
\begin{figure}[h]
    \centering
    \includegraphics[width=\linewidth]{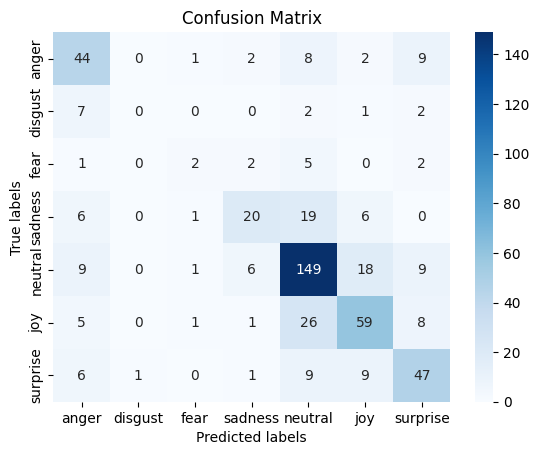}
    \caption{Confusion matrix for emotion recognition with context using fine-tuned Llama}
    \label{fig:cm3}
\end{figure}
\begin{figure}[h]
    \centering
    \includegraphics[width=\linewidth]{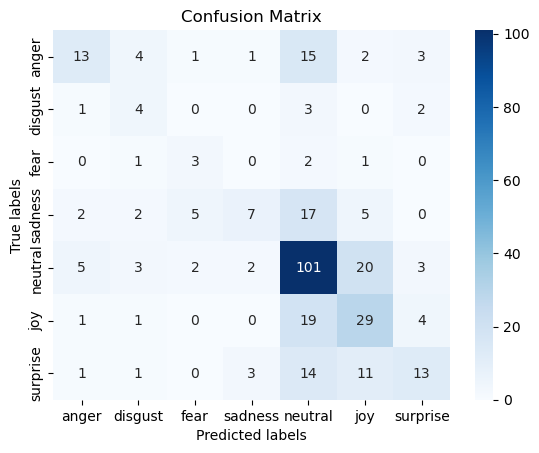}
    \caption{Confusion matrix for emotion recognition using Zero-shot GPT-3.5}
    \label{fig:zero-shot-cm}
\end{figure}
\begin{figure}[h]
    \centering
    \includegraphics[width=\linewidth]{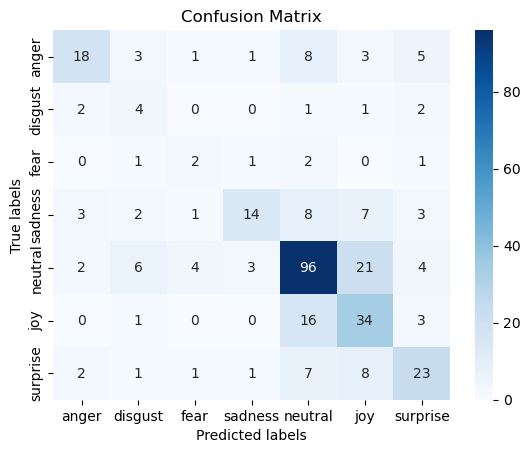}
    \caption{Confusion matrix for emotion recognition using GPT-ICL}
    \label{fig:gpt-icl-cm}
\end{figure}

\begin{table*}[htbp]
  \centering
  \small{
  % \resizebox{\textwidth}{!}{%
  \begin{tabular}{llccccccccc}
    \hline
    \multirow{2}{*}{\textbf{Approach}} & \textbf{Metric} & \textbf{Anger} & \textbf{Disgust} & \textbf{Fear} & \textbf{Joy}
    & \textbf{Sadness} & \textbf{Surprise} & \textbf{Neutral} \\
     & \textbf{Supp} & 71 & 13 & 13 & 101 & 54 & 77 & 209 \\
    \hline
    \multirow{3}{*}{\textbf{Zero-shot Llama w/o context}} & \textbf{P} & 0.0000 & 0.0000 & 0.0000 & 0.1881 & 0.0000 & 0.0000 & 0.0000 \\
    & \textbf{R} & 0.0000 & 0.0000 & 0.0000 & 1.0000 & 0.0000 & 0.0000 & 0.0000 \\
    & \textbf{F1} & 0.0000 & 0.0000 & 0.0000 & 0.3166 & 0.0000 & 0.0000 & 0.0000 \\
    \hline
    \multirow{3}{*}{\textbf{Zero-shot Llama with context}} & \textbf{P} & 0.7500 & 0.2857 & 0.0000 & 0.3798 & 0.4583 & 0.3200 & 0.5663 \\
    & \textbf{R} & 0.0845 & 0.1538 & 0.0000 & 0.7822 & 0.2037 & 0.5195 & 0.4498 \\
    & \textbf{F1} & 0.1519 & \textbf{0.2000} & 0.0000 & 0.5113 & 0.2821 & 0.3960 & 0.5013 \\
    \hline
    \multirow{3}{*}{\textbf{Fine-tuned Llama with context}} & \textbf{P} & 0.5641 & 0.0 & 0.3333 & 0.6210 & 0.625 & 0.6103 & 0.6666 \\
    & \textbf{R} & 0.6197 & 0.0 & 0.1538 & 0.5842 & 0.3704 & 0.6104 & 0.7943 \\
    & \textbf{F1} & \textbf{0.5906} & 0.0 & \textbf{0.2105} & \textbf{0.6020} & \textbf{0.4651} & \textbf{0.6104} & \textbf{0.7249} \\
    \hline
    \multirow{3}{*}{\textbf{Zero-Shot GPT}} & \textbf{P} & 0.5652 & 0.2500 & 0.2727 & 0.4265 & 0.5385 & 0.5200 & 0.5906 \\
    & \textbf{R} & 0.3333 & 0.4000 & 0.4286 & 0.5370 & 0.1842 & 0.3023 & 0.7426 \\
    & \textbf{F1} & 0.4194 & 0.3077 & 0.3333 & 0.4754 & 0.2745 & 0.3824 & 0.6580 \\
    \hline
    \multirow{3}{*}{\textbf{In-Context-Learning GPT}} & \textbf{P} & 0.6667 & 0.2222 & 0.2222 & 0.4595 & 0.7000 & 0.5610 & 0.6957 \\
    & \textbf{R} & 0.4615 & \textbf{0.4000} & \textbf{0.2857} & 0.6296 & 0.3684 & 0.5349 & 0.7059 \\
    & \textbf{F1} & \textbf{0.5455} & 0.2857 & 0.2500 & \textbf{0.5312} & \textbf{0.4828} & \textbf{0.5476} & \textbf{0.7007} \\
    \hline
  \end{tabular}
  }
  \caption{Emotion Recognition Results for Seven Emotion Categories. P: precision, R: recall, F1: F1 score,  and Supp: support.}
  \label{tab:emotion_results}
\end{table*}

\section{Prompt details}
\label{appendix:prompts}
\subsection{Fine-tuned Llama 2}
\label{appendix:llama-prompts}
The general prompt for the Llama chat version is given in Figure \ref{fig:general_prompt}. The prompts for emotion and cause prediction are given in Fig.~\ref{fig:emotion_prompt1} and Fig.~\ref{fig:cause_prompt1}. We provide a specific format for the output so as to ease the post-processing where we extract the first emotion label occurring after the "::" sequence of characters.

\begin{figure}[htbp]
    \centering
    \includegraphics[width=\linewidth]{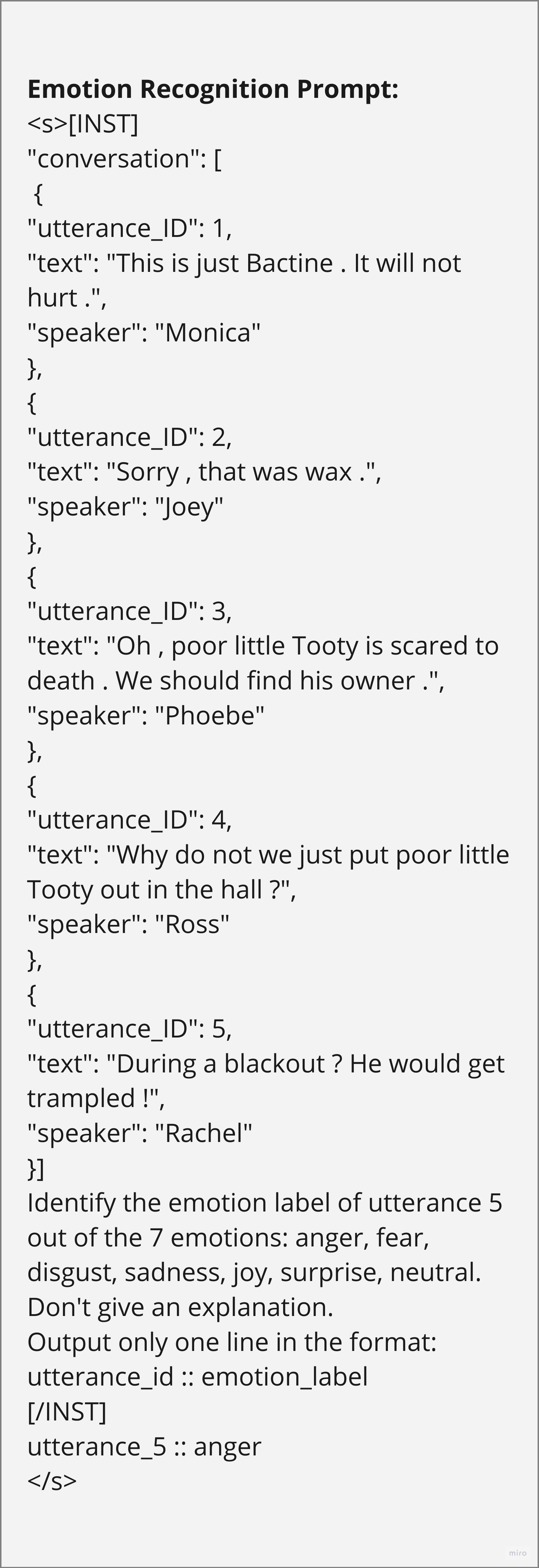}
    \caption{Example Prompt for emotion prediction using Llama}
    \label{fig:emotion_prompt1}
\end{figure}
\begin{figure}[htbp]
    \centering
    \includegraphics[width=\linewidth]{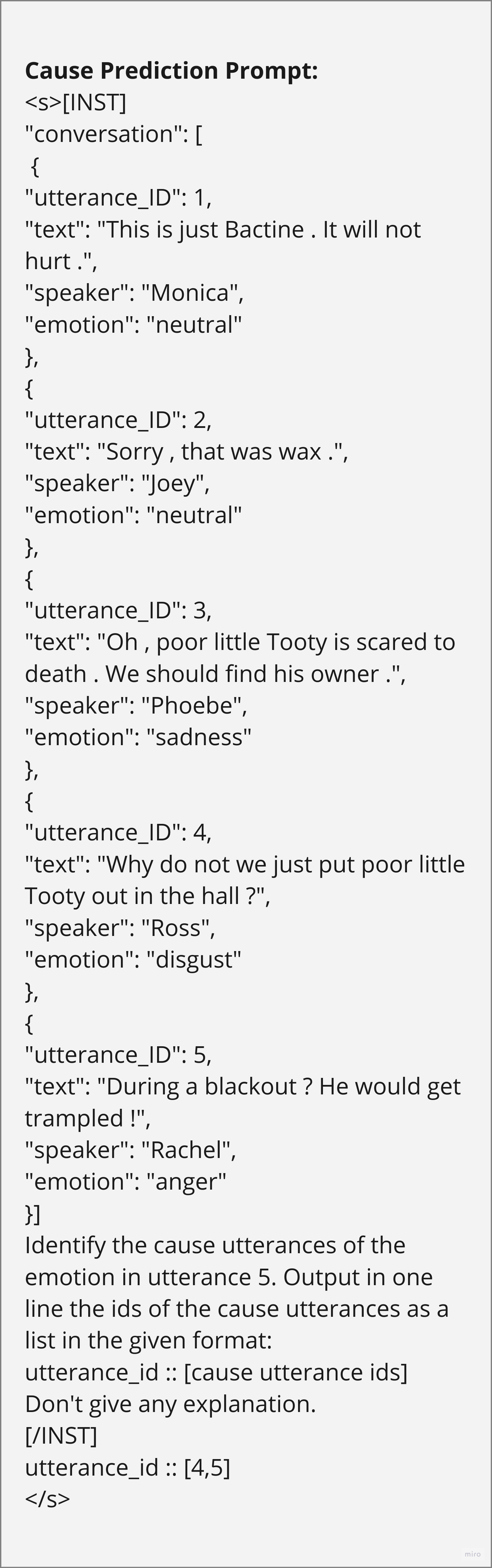}
    \caption{Example Prompt for cause prediction using Llama}
    \label{fig:cause_prompt1}
\end{figure}
\begin{figure}[htbp]
    \centering
    \includegraphics[width=0.85\linewidth]{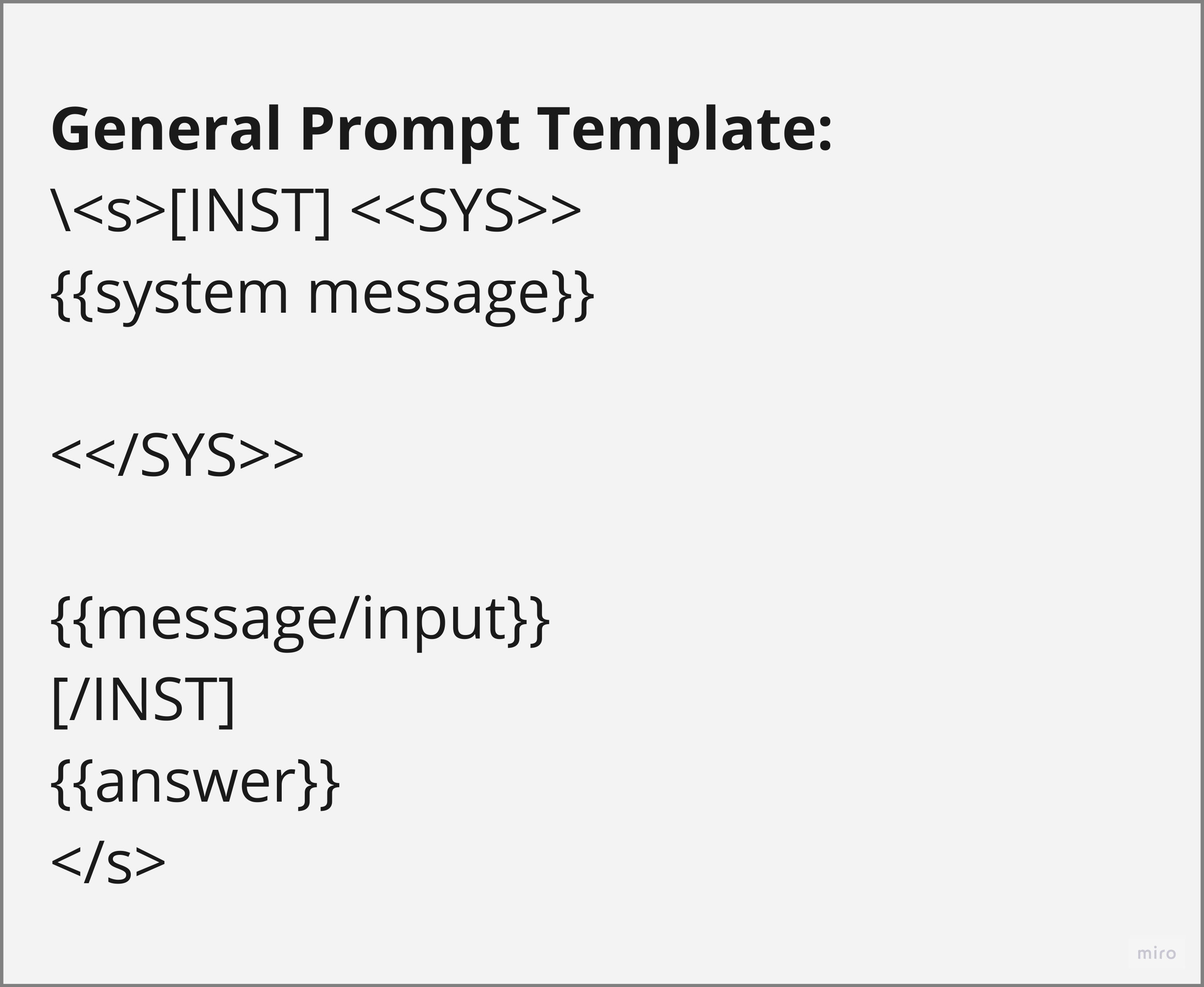}
    \caption{General Prompt Template for Llama}
    \label{fig:general_prompt}
\end{figure}

\subsection{ICL-GPT}
\label{appendix:gpt-prompts}
We devise prompt templates to be used in the LangChain framework. \{\} represent placeholders to be replaced when making a prompt. Video captioning prompt is given in Fig.~\ref{fig:video-caption-prompt}. Due to rate limits, we had to batch the utterances, thus we may have multiple disjoint descriptions of a conversation. We prompt GPT-3.5 using the prompt in Fig.~\ref{fig:batched-caption-stitching} to stitch the descriptions into a single caption. For explaining the retrieved conversation with emotion annotated, we use the prompt in Fig.~\ref{fig:emotion-explanation}. The retrieved conversation and explanation are now used as demonstration examples for the emotion recognition prompt in Fig.~\ref{fig:erc-context-prompt}. For an explanation of causes in the retrieved-context window, we use the prompt in Fig.~\ref{fig:cause-explanation}. The explanations of the retrieved windows are used as demonstration examples in the prompt for cause prediction within a context window as shown in the prompt in Fig.~\ref{fig:cpred-context-prompt}.

\begin{figure}[h!]
    \centering
    \includegraphics[width=\linewidth]{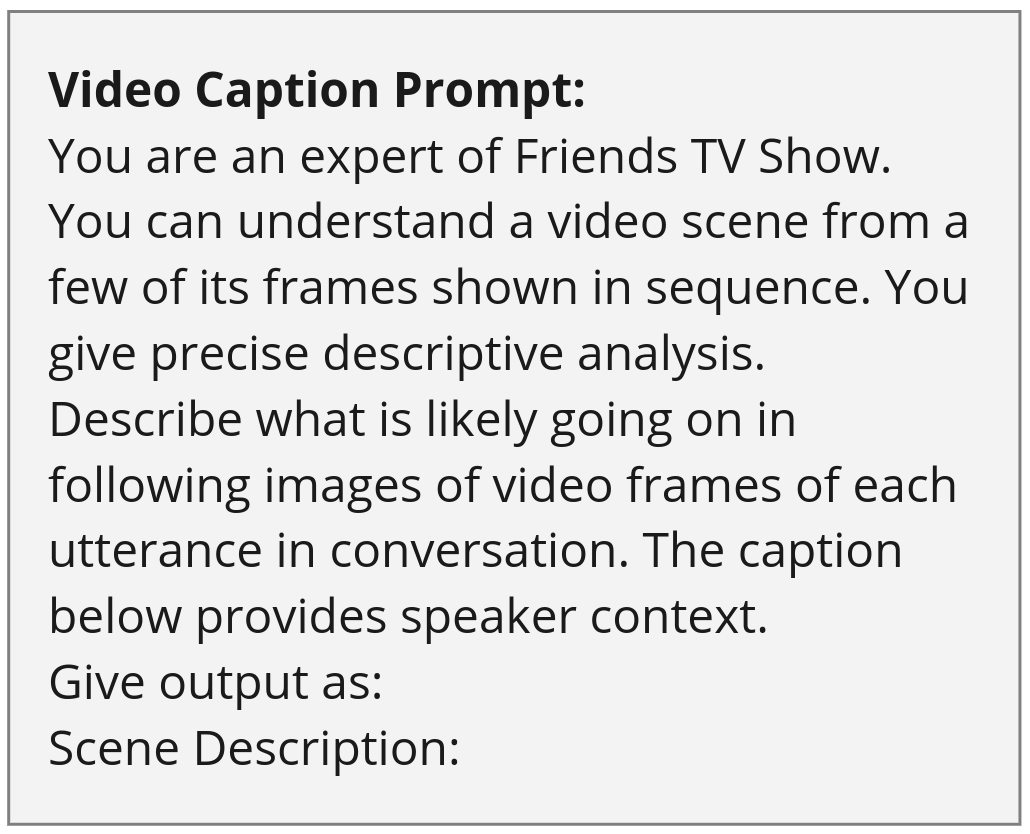}
    \caption{Video Captioning Prompt Template}
    \label{fig:video-caption-prompt}
\end{figure}
\begin{figure}[t]
    \centering
    \includegraphics[width=\linewidth]{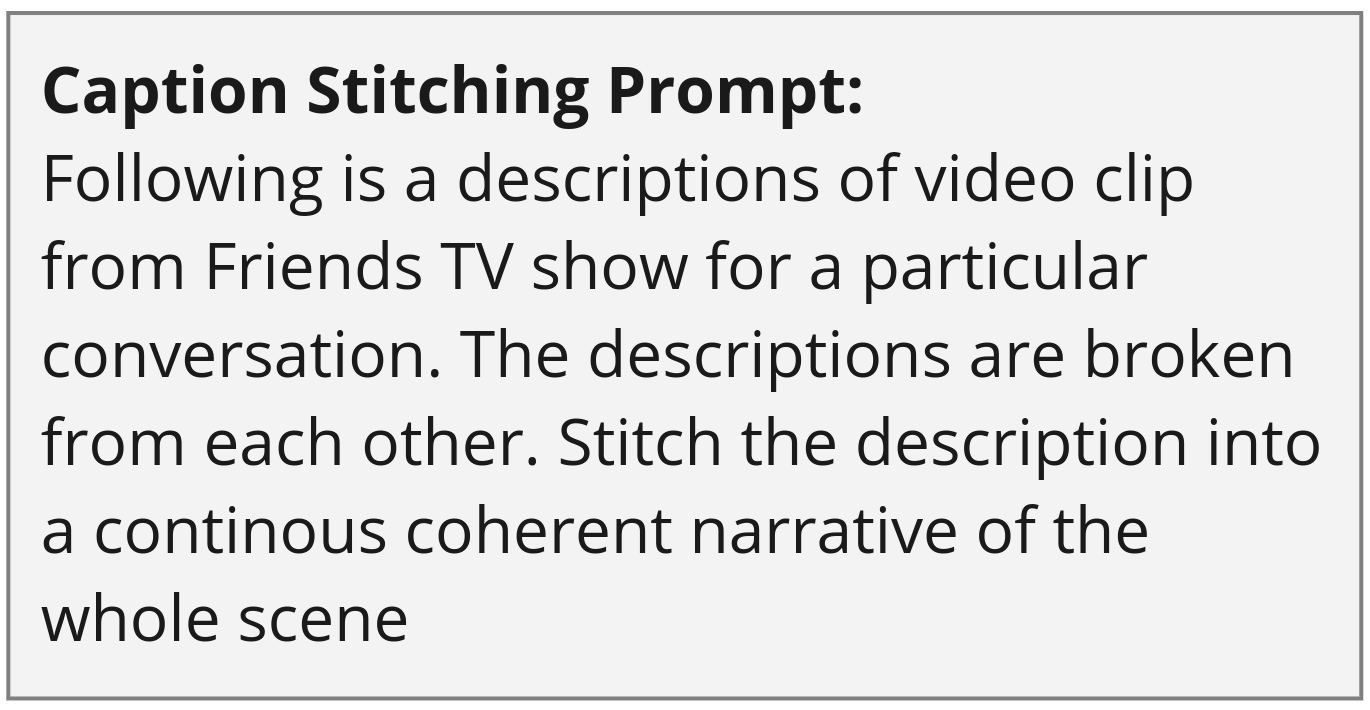}
    \caption{Batched Video Caption Stitching Prompt Template}
    \label{fig:batched-caption-stitching}
\end{figure}
\begin{figure}[t]
    \centering
    \includegraphics[width=\linewidth]{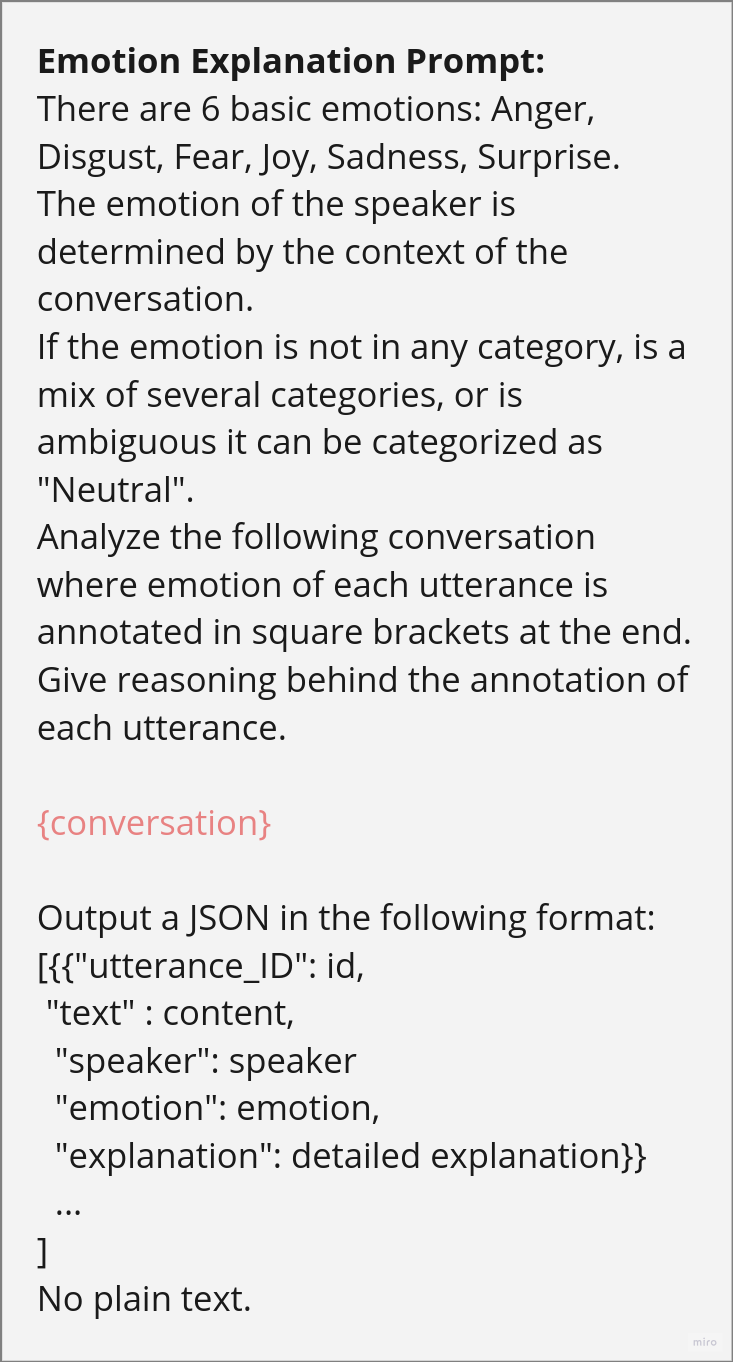}
    \caption{Emotion Label Explanation Prompt Template}
    \label{fig:emotion-explanation}
\end{figure}
\begin{figure}[t]
    \centering
    \includegraphics[width=\linewidth]{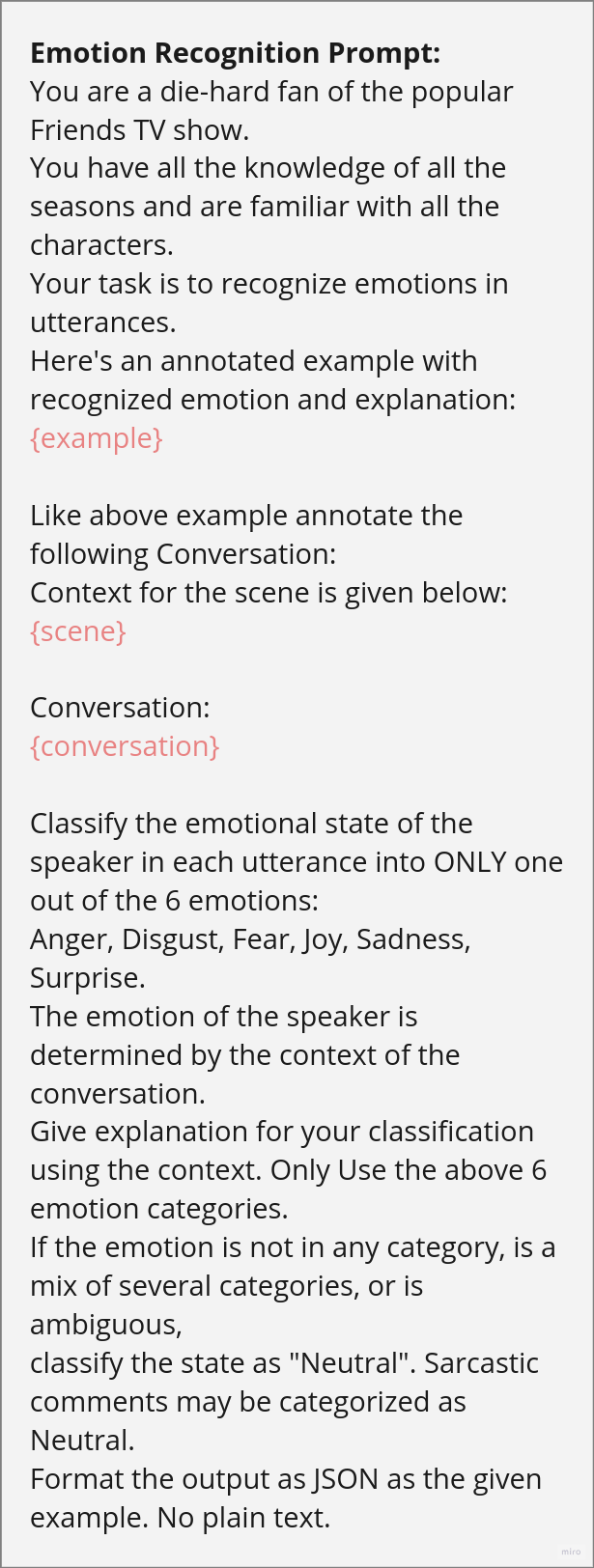}
    \caption{Emotion Recognition with Context Learning Prompt Template}
    \label{fig:erc-context-prompt}
\end{figure}
\begin{figure}[t]
    \centering
    \includegraphics[width=\linewidth]{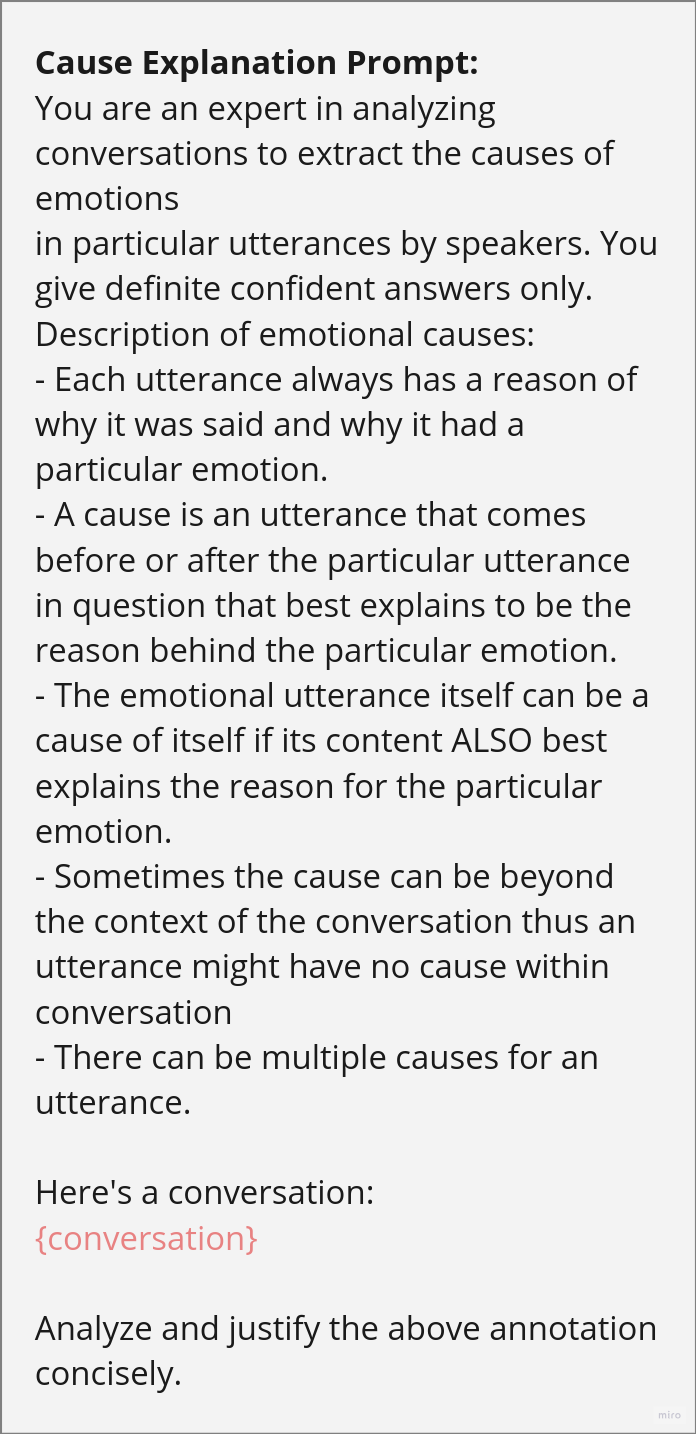}
    \caption{Cause Explanations Prompt Template}
    \label{fig:cause-explanation}
\end{figure}
\begin{figure}[t]
    \centering
    \includegraphics[width=\linewidth]{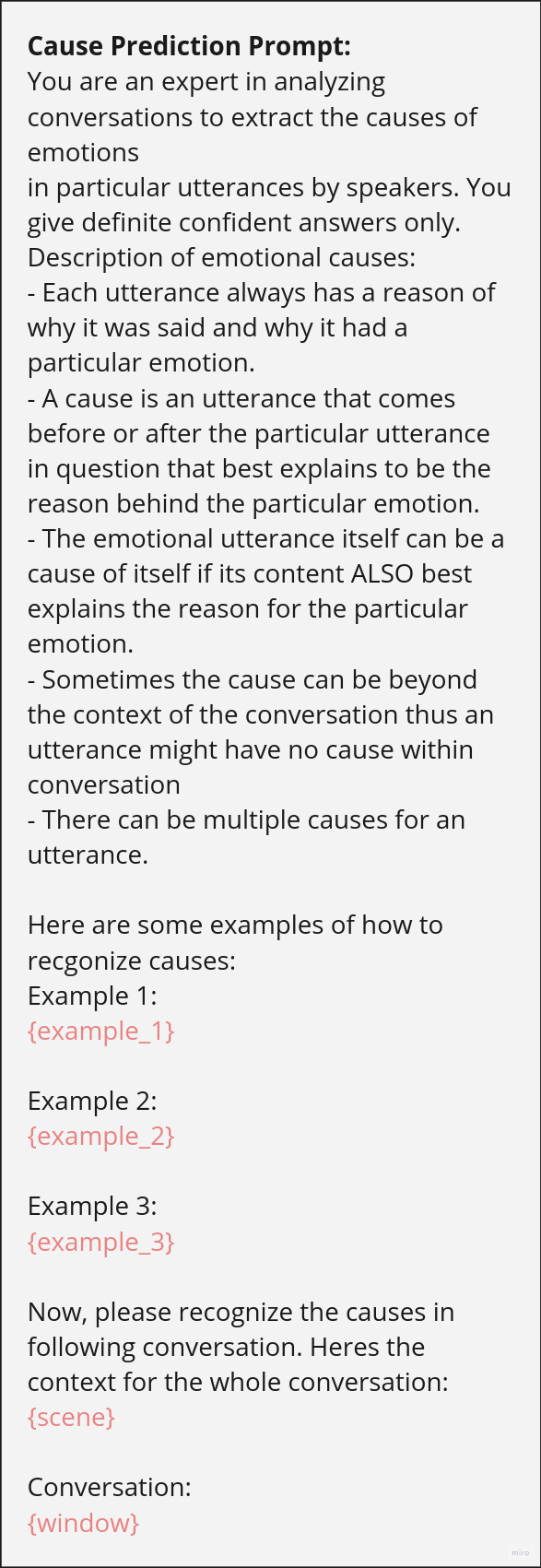}
    \caption{Cause Prediction with Context Learning Prompt Template}
    \label{fig:cpred-context-prompt}
\end{figure}

\end{document}